\def\year{2021}\relax
\documentclass[letterpaper]{article} 
\usepackage{aaai21}  
\usepackage{times}  
\usepackage{helvet} 
\usepackage{courier}  
\usepackage[hyphens]{url}  
\usepackage{graphicx} 
\usepackage{natbib}  
\usepackage{caption} 
\usepackage{hyperref}

\usepackage{url}            
\usepackage{booktabs}       
\usepackage{amsfonts}       
\usepackage{nicefrac}       
\usepackage{microtype}      
\usepackage{graphicx}
\usepackage{subfigure}
\usepackage{amsmath}
\usepackage{amssymb}
\usepackage{amsthm}
\usepackage{dsfont}
\usepackage{multirow}
\usepackage{bbding}
\usepackage{caption}
\usepackage{tabularx}
\usepackage{pifont}
\usepackage{color}
\usepackage{arydshln}
\usepackage[ruled, vlined]{algorithm2e}

\SetKwInput{KwGiven}{Define}
\SetKwInput{KwGiven}{Given}
\SetKwInput{KwHparams}{Hyperparams}

\newcommand{\eos}{\ensuremath{\left<\text{eos}\right>}}

\defcitealias{wang2020exposure}{W \& S (2020)}
\defcitealias{edunov2018classical}{Ed. (2018)}

\title{MLE-Guided Parameter Search for Task Loss Minimization in Neural Sequence Modeling}
\author{
    Sean Welleck,\thanks{Correspondence to: wellecks@nyu.edu.}
    Kyunghyun Cho\textsuperscript{\rm 1}\\
}
\affiliations{
    New York University\\
    \textsuperscript{\rm 1}CIFAR Associate Fellow\\
}

\begin{document}

\maketitle

\begin{abstract}
Neural autoregressive sequence models are used to generate sequences in a variety of natural language processing (NLP) tasks, where they are evaluated according to sequence-level task losses.
These models are typically trained with maximum likelihood estimation, which ignores the task loss, yet empirically performs well as a surrogate objective.
Typical approaches to directly optimizing the task loss such as policy gradient and minimum risk training are based around sampling in the sequence space to obtain candidate update directions that are scored based on the loss of a single sequence.
In this paper, we develop an alternative method based on random search in the parameter space that leverages access to the maximum likelihood gradient.
We propose maximum likelihood guided parameter search (MGS), which samples from a distribution over update directions that is a mixture of random search around the current parameters and around the maximum likelihood gradient, with each direction weighted by its improvement in the task loss.
MGS shifts sampling to the parameter space, and scores candidates using losses that are pooled from multiple sequences.
Our experiments show that MGS is capable of optimizing sequence-level losses, with substantial reductions in repetition and non-termination in sequence completion, and similar improvements to those of minimum risk training in machine translation.
\end{abstract}

\section{Introduction}

Neural autoregressive sequence models are used in a variety of natural language processing (NLP) tasks, such as machine translation \citep{bahdanau2014neural}, summarization \citep{rush2015}, dialogue modeling \citep{vinyals2015neural}, and text completion \citep{sutskever2011generating,graves2013generating,radford2018gpt2,holtzman2019curious,welleck2020neural}. 
In these tasks, a decoding algorithm is used to produce sequences that are evaluated according to a sequence (or corpus) level task loss such as \textsc{Bleu} \citep{papineni2002bleu}, \textsc{Meteor} \citep{banerjee2005meteor}, or alternative n-gram based metrics.

The conventional training approach, maximum likelihood, optimizes a token-level surrogate to the 0-1 loss, and only leverages sequences drawn from the ground-truth distribution.
The resulting mismatch between the training and evaluation loss functions, and the discrepancy between the sequence distributions used for training and the distribution encountered at evaluation time has prompted alternative sequence-level training algorithms (e.g. \cite{daume2009searn,ranzato2016mixer,shen2016mrt}). 
Nevertheless, maximizing the likelihood has empirically performed well as a surrogate to minimizing the task loss, achieving strong performance on the aforementioned tasks \citep{ott2019scaling,raffel2019}.
In this paper, we develop a sequence-level training procedure that addresses the downsides of maximum likelihood by leveraging its strengths as a surrogate objective.

It is challenging to optimize a task loss, as the loss is typically non-differentiable with respect to the model parameters, and optimization is done over a high-dimensional parameter space.
Typical approaches to this problem in natural language processing are based around the policy gradient estimator \citep{williams1992}, such as  \citet{shen2016mrt,ranzato2016mixer,bahdanau2016actorcritic,yu2017seqgan}.
This estimator is used to optimize an arbitrary task loss by introducing stochasticity via autoregressive sampling in the action space, which is a critical downside in NLP, where the action space (vocabulary) is large and the sequence-level reward is sparse.
The estimator's variance grows with the sequence length, necessitating a parameterized baseline 
or a heuristic sampling schedule in practice, while requiring initialization from a pre-trained model.
Recently, the effectiveness of these methods in NLP has been called into question \citep{caccia2020language,choshen2020weaknesses}.

An alternative class of 
methods optimize a black-box function without requiring gradient information.
Of these, estimation-of-distribution algorithms, including the cross-entropy method \citep{rubinstein1999,deBoer2005},
evolutionary strategies \citep{rechenberg1978,back1993overview},
and their variants \citep{hansen2001cmaes,wierstra2014natural,salimans2017evolution}, operate by maintaining a search distribution from which a set of random perturbations are sampled. 
The function value (i.e. task loss) at each of the perturbed points is used to update the search distribution.
Because stochasticity is introduced in the parameter space rather than the action space, rewards associated with each search candidate are pooled over multiple examples, `densifying' the sparse reward.
This parameter-space exploration \citep{ruckstie2010exploring,plappert2018parameter} 
is attractive for NLP since the same decoding algorithm can be used for training and evaluation, and the variance is independent of the action space size or the sequence length.
However, a key challenge for these black-box methods is handling high-dimensional search spaces.
This typically restricts their use to training networks that are small compared to those used in neural sequence modeling \citep{mania2018simple,ha2018world},
or requires massive parallelization \citep{salimans2017evolution}, and their use with large-scale natural language processing models has been under-explored.

In this paper, we leverage the fact that in many sequence modeling tasks encountered in natural language processing, a surrogate update direction is available in the form of the maximum likelihood gradient.
We hypothesize that incorporating this surrogate information into a random-search method can substantially alleviate issues stemming from the large search space.
We frame learning as sampling from a distribution over parameter update directions that is proportional to the improvement in task loss.
Since this distribution is only accessible for evaluation up to a normalizing constant, we propose to use self-normalized importance sampling for obtaining update directions.

The key idea behind our method is to form a proposal distribution that is a mixture of random search around the current parameters and around the maximum-likelihood update direction.
Our experiments show that the resulting procedure, called \textit{maximum-likelihood guided parameter search} (MGS), is effective for minimizing sequence-level losses in natural language generation and machine translation, offering an alternative to policy gradient and minimum risk methods.

\section{Maximum-Likelihood Guided Parameter Search}
\paragraph{Sequence generation.}
Sequence generation is the problem of mapping an input $X$ to an output $Y=(y_1,\ldots,y_{|Y|})$.
In our setting of neural sequence generation, this mapping is a deterministic decoding algorithm $\mathcal{F}(\theta, X)$, which uses an autoregressive model
$p_{\theta}(Y|X)=\prod_{t=1}^{|Y|}p_{\theta}(y_t|y_{<t}, X)$ to produce an output $\hat{Y}$ given an input $X$.
This includes greedy and beam search decoding, and stochastic decoding algorithms with a noise input, $\mathcal{F}(\theta, X, \epsilon)$.
The goal of sequence generation is to find a model whose generations have minimal task loss on a set $\mathcal{D}=\{(X,Y)\}$ of input-output pairs,
\begin{align}
\label{eqn:seqgen-objective}
    \displaystyle C(\theta, \mathcal{D}) = \sum_{X,Y\in \mathcal{D}}c(\mathcal{F}(\theta, X), Y),
\end{align}
where we assume $c(\hat{Y}, Y)\in \mathbb{R}$ is an arbitrary sequence-level loss (e.g. sentence-\textsc{Bleu}).
The most widely used approach to training such a model is minimizing the negative log-likelihood given a training set, which ignores the task loss : $\mathcal{L}_{\text{MLE}}(\theta; \mathcal{D}) = - \sum_{X,Y\in \mathcal{D}}\sum_{t=1}^{|Y|}\log p_{\theta}(y_t|y_{<t}, X)$.

\paragraph{Method.}
To directly optimize (\ref{eqn:seqgen-objective}), we iteratively update the parameters $\theta$ in the direction of maximal improvement in the task loss. 
Each update corresponds to the expected update under a distribution that weights each direction according to its improvement,
\begin{align}
    \displaystyle \Delta_* &= 
    \mathbb{E}_{\Delta\sim p_*(\Delta|\theta;\alpha)}\left[\Delta\right],
    \label{eqn:ggs-ptilde}
\end{align}
where,
\begin{align*}
    p_*(\Delta|\theta;\alpha) \propto \tilde{p}_*(\Delta|\theta;\alpha) =\exp \left(\alpha(C(\theta)-C(\theta+\Delta))\right),
\end{align*}
and 
$\alpha\in \mathbb{R}_{>0}$ is a temperature parameter.
When $\alpha\rightarrow 0$, the distribution becomes uniform, and when $\alpha\rightarrow\infty$ it concentrates on the direction(s) of maximal task loss improvement.
Since $p_*$ is only known up to a normalizing constant and is defined over a high-dimensional parameter space, it is impractical to approximate the update direction $\Delta_*$ with samples from $p_*$.
Instead, we use self-normalized importance sampling with a proposal distribution $q(\Delta|\theta)$:
\begin{align}
\label{eqn:update}
    \displaystyle \Delta_* &= 
    \mathop{\mathbb{E}}_{\Delta\sim q(\Delta|\theta)} \left[\frac{p_*(\Delta|\theta;\alpha)}{q(\Delta|\theta)}\Delta\right]\\
    & \approx \sum_{k=1}^K\frac{ w(\Delta_k)}{\sum_{k=1}^K w(\Delta_k)}\Delta_k = \Delta_{\text{MGS}},
\end{align}
where 
\begin{align}
\label{eqn:weight}
w(\Delta_k) &=
\frac{\exp \left(\alpha(C(\theta)-C(\theta+\Delta_k))\right)}{\tilde{q}(\Delta_k|\theta)},
\end{align}
$\Delta_k\sim q(\Delta|\theta)$, 
and $q\propto \tilde{q}$.
This update direction equals $\Delta_*$ in the limit:  $\mathbb{P}\left(\lim_{K\rightarrow\infty}\Delta_{\text{MGS}}=\Delta_*\right)=1$ \citep{owen2013monte}.\footnote{See Appendix~\ref{apx:snis} for a review of self-normalized importance sampling.}

The sample complexity of such a random-search method is known to depend on the dimensionality of the sample space \citep{vemula19contrasting}, thus it is crucial to choose a good proposal distribution. 
Our contribution is a proposal distribution for use in sequence generation, where we have access to the maximum likelihood gradient $\nabla_\theta \mathcal{L}_{\text{MLE}}$.
Specifically, we propose a mixture of two Gaussians, whose components are centered at the origin and at the maximum-likelihood gradient, respectively:
\begin{align}
\label{eqn:proposal}
    q_{\text{MGS}}(\Delta|\theta) = &\ \mathcal{N}(\Delta|0, I\sigma^2)\cdot \pi  + \\
    \nonumber
    &\ \mathcal{N}(\Delta|\nabla_{\theta}\mathcal{L}_{\text{MLE}}, I\sigma^2)\cdot (1-\pi),
\end{align}
where $\pi\in [0,1]$ is a mixture parameter that we set to 0.5 in practice.
Given a batch of examples, we compute the gradient of the maximum likelihood loss, sample candidate directions from the proposal distribution (\ref{eqn:proposal}), then evaluate the task loss of each candidate and form the update direction (\ref{eqn:update}).
Algorithm \ref{alg:mgs} summarizes the procedure, called \textbf{maximum-likelihood guided parameter search} (MGS).

\begin{algorithm}[t]
\DontPrintSemicolon
\caption{MLE-guided parameter search (MGS).}
\label{alg:mgs}
\KwGiven{Batch $\{X_i, Y_i\}_{i=1}^{B}$, model $p_{\theta}$, decoding algorithm $\mathcal{F}$, task-loss $c(\hat{Y}, Y)$.}
\KwHparams{Number of candidates $K$, temperature $\alpha$, noise level $\sigma^2$.}
\KwOut{Update direction $\Delta_{\text{MGS}}$.}
$\{\hat{Y}_i\}=\mathcal{F}(\theta, \{X_i\})$ \tcp*[r]{decode}
$C(\theta) = \frac{1}{B}\sum_{i=1}^B c(\hat{Y}_i, Y_i)$\tcp*[r]{eqn. \ref{eqn:seqgen-objective}} 
$\nabla_{\theta}\mathcal{L}_{\text{MLE}} = \texttt{backprop}(\mathcal{L}_{\text{MLE}}(\theta; \{X_i, Y_i\}))$\\
\For{$k\in1,\ldots,K$}{
    $\Delta_k\sim q_{\text{MGS}}(\cdot|\theta,\nabla_{\theta}\mathcal{L}_{\text{MLE}},\sigma^2)$\tcp*[r]{eqn. \ref{eqn:proposal}} 
    $\{\hat{Y}_i\}=\mathcal{F}(\theta+\Delta_k, \{X_i\})$\tcp*[r]{decode}
    $C(\theta+\Delta_k) = \frac{1}{B}\sum_{i=1}^B c(\hat{Y}_i, Y_i)$\tcp*[r]{eqn. \ref{eqn:seqgen-objective}} 
    $w(\Delta_k)=\frac{\exp\left(\alpha(C(\theta)-C(\theta+\Delta_k))\right)}{q_{\text{MGS}}(\Delta_k|\theta)}$ \tcp*[r]{eqn. \ref{eqn:weight}} 
}
$\Delta_{\text{MGS}}=\sum_{i=1}^{K}\frac{w(\Delta_k)}{\sum_{k'}w(\Delta_{k'})}\Delta_k$\tcp*[r]{eqn. \ref{eqn:update}}
\end{algorithm}
\section{Other Task Loss Minimization Methods}
\label{sec:comparison}
\paragraph{Comparison with policy gradient.}
Policy gradient (PG) methods such as REINFORCE \citep{williams1992} consist of the objective and gradient estimator:
\begin{align}
    \label{eqn:pg-cost}
    \displaystyle C_{\text{PG}}(\theta) = \mathop{\mathbb{E}}_{
    (X,Y)\sim \mathcal{D}
    }
    \mathbb{E}_{
    \hat{Y}\sim p_{\theta}(\cdot|X)}
    \left[c(\hat{Y}, Y)\right], \\
    \displaystyle\nabla^{\text{PG}}_{\theta} = \mathbb{E}_{\hat{Y}\sim p_{\theta}(\cdot|X)}\left[c(\hat{Y}, Y)\nabla_\theta \log p_{\theta}(\hat{Y}|X)\right].
\end{align}
The policy gradient objective contains an expectation over the output distribution $p_{\theta}(\cdot|X)$, unlike the objective optimized by MGS (Equation \ref{eqn:seqgen-objective}).
In particular, computing the PG objective involves decoding with ancestral sampling, while the objective (\ref{eqn:seqgen-objective}) uses an arbitrary decoding algorithm.
Naturally, approximating the policy gradient also uses ancestral sampling instead of the algorithm used at inference time (e.g. greedy or beam search).
To contrast this with maximum-likelihood guided parameter search, we formalize the sampling and examine the per-sequence gradient.

Ancestral sampling decodes a sequence by sampling auto-regressively from the model's per-step categorical distributions. 
Given noise $\epsilon\sim \mathcal{U}(0,1)$, ancestral sampling, which consists of repeated categorical sampling $\hat{y}_t\sim p_{\theta}(\cdot|\hat{y}_{<t},X)$, can be written as a deterministic function $\hat{Y}=\mathcal{F}_{\text{anc}}(\theta, X, \epsilon)$.
The policy gradient estimator is an expectation over the noise used to produce the categorical samples,
\begin{align*}
    \displaystyle\nabla^{\text{PG}}_{\theta} &= 
    \mathbb{E}_{\epsilon}\left[c\left(\mathcal{F}_{\text{anc}}(\theta, X, \epsilon), Y\right)\nabla_\theta \log p_{\theta}(\mathcal{F}_{\text{anc}}( \theta, X, \epsilon))\right].
\end{align*}
Maximum-likelihood guided parameter search uses any arbitrary decoding algorithm, e.g. $\hat{Y}=\mathcal{F}_{\text{greedy}}(\theta, X)$, which can be chosen to be the same algorithm used at evaluation time.
The MGS estimator is an expectation over noise in the \textit{parameter space}, $\nabla_{\theta}^{\text{MGS}}=$
\begin{align*}
    \mathbb{E}_{\epsilon\sim q}\left[\hat{w}(\epsilon)\exp\left(\alpha(c(\mathcal{F}(\theta, X), Y) - c(\mathcal{F}(\theta+\epsilon, X), Y)\right)\epsilon\right],
\end{align*}
where we consider a single example and rewrite the MGS update (\ref{eqn:update}) 
in order to illustrate how the use of noise and the decoding algorithm differ from policy gradient.
See the Appendix for the derivation.
In short, policy gradient uses each parameter $\theta$ to sample multiple sequences for each input, while MGS samples multiple parameters, and uses each to decode a single sequence per input.

\paragraph{Comparison with minimum risk training.}
Minimum risk training (MRT) \citep{shen2016mrt} approximates the policy gradient objective ($\ref{eqn:pg-cost}$) as,
\begin{align}
    \label{eqn:mrt-cost}
    \displaystyle C_{\text{MRT}}(\theta) &= \mathop{\mathbb{E}}_{\substack{(X,Y)\sim \mathcal{D}}} 
    \mathop{\mathbb{E}_{\hat{Y}\sim q_{\theta}(\cdot|X, S)}}\left[c(\hat{Y}, Y)\right],\\
    q_{\theta}(Y|X, S) &= \begin{cases}
        \frac{p_{\theta}(Y|X)^\alpha}{Z_{\theta}(X,S)}, & \text{if }Y\in S,\\
        0,         & \text{otherwise},
    \end{cases}
\end{align}
where $S=\{\hat{Y}_1,\ldots,\hat{Y}_k\}$ is a set of candidate output sequences, and $Z_{\theta}(X,S)=\sum_{Y\in S}p_{\theta}(Y|X)^\alpha$.
There are no importance weights, and $q_{\theta}$ is not a valid proposal, unlike $q_{\text{MGS}}$.
The gradient is,\footnote{See Appendix~\ref{apx:mrt-derivation} for the derivation.}
\begin{align}
    \label{eqn:grad-mrt}
    \displaystyle \nabla_{\theta}C_{\text{MRT}} &= 
    \alpha\left[
    \mathbb{E}_{q_{\theta}}\left[c(\hat{Y},Y)\nabla_\theta\log p_\theta(\hat{Y}|X)\right]-\right.\\
    \nonumber
    &\quad\quad\ \left. \mathbb{E}_{q_{\theta}}\left[c(\hat{Y},Y)\right]
    \mathbb{E}_{q_{\theta}}\left[\nabla_\theta\log p_\theta(\hat{Y}|X)\right]\right],
\end{align}
where $\mathbb{E}_{q_{\theta}}$ denotes $\mathbb{E}_{\hat{Y}\sim q_{\theta}(\cdot|X, S)}$. 
The MRT gradient consists of the policy gradient, minus a term that includes the score function and the expected loss.
Minimum risk training can incorporate the maximum likelihood gradient by including the ground truth sequence $Y^*$ as a candidate,
\begin{align*}
    \nabla_{\theta}C_{\text{MRT}} &= 
        \alpha[
          \left(w(Y^*)-\bar{w}(Y^*)\right)
          \nabla_\theta\log p_\theta(Y^*|X)
          + \\
          & \quad\quad\sum_{\hat{Y}\in S\backslash Y^*}\left(w(\hat{Y})-\bar{w}(\hat{Y})\right)\nabla_\theta\log p_\theta(\hat{Y}|X)
          ]
\end{align*}
where $w(Y')=c(Y', Y)q_{\theta}(Y'|X, S)$, and $\bar{w}(Y')=\mathbb{E}_{Y''\sim q_\theta}\left[c(Y'',Y)\right]q_{\theta}(Y'|X,S)$.
Unlike MGS, the other candidate directions in MRT are not related to the maximum-likelihood gradient.
Instead, the candidates are determined by action-space sampling, similar to policy gradient.

\paragraph{Pooled task losses.} PG and MRT both sample in the action space (i.e. vocabulary), while  the proposed MGS samples in the parameter space. 
This difference affects the \textit{amount of supervision} that is used to weight each candidate update direction.
To see this, consider a minibatch $\{X_n, Y_n\}_{n=1}^N$.
The policy gradient estimator with $K$ samples per batch element is,
\begin{align}
    \nabla^{\text{PG}}_{\theta} &= \frac{1}{NK}\sum_{n,k}c(\hat{Y}_{n}^{(k)}, Y_n)\nabla_{\theta}\log p_{\theta}(\hat{Y}_{n}^{(k)}|X_n),
\end{align}
where $\hat{Y}^{(k)}_n$ is a sampled sequence.
Policy gradient uses a \textit{single} sequence loss to weight each candidate update direction.
A similar inspection reveals that MRT shares this property.
On the other hand, MLE-guided parameter search,
\begin{align*}
    \nabla_{\theta}^{\text{MGS}} &= \sum_{k}\left[\hat{w}(\Delta_k)\exp\left(\alpha(C(\theta) - C(\theta+\Delta_k))\right)\Delta_k\right],
\end{align*}
weights each candidate direction using a loss $C(\cdot)$ computed over the entire minibatch (see Equation \ref{eqn:seqgen-objective}).
This has the effect of `densifying' the sparse loss by pooling the losses from multiple examples.

\begin{table*}
\centering
\small 
\begin{tabular}{lllllll}
& \textbf{LM} $\downarrow$ & \textbf{Edit} $\downarrow$ & \textbf{Nonterm} $\downarrow$ & \textbf{Repetition} $\downarrow$& \textbf{Avg. len.} & \textbf{Perplexity}$\downarrow$\\ 
\toprule
\textbf{MLE} & 157.6 (13.5) & .945 (.008) & .344 (.063) & .530 (.062) & 228.1 (33.3) & 21.3 (0.2)\\
\midrule
\textbf{MGS-LM}           & 64.9 (2.09) & .937 (.002) & .012 (.003) & .046 (.009) & 22.8 (2.2) & 22.0 (0.1) \\ 
\textbf{MRT-LM (+MLE 0.1)}   & 57.4 (.967) & .948 (.002) & .013 (.004) & .023 (.005) & 16.9 (2.3) & 25.8 (1.7) \\ 
\textbf{PG-LM (+MLE 0.1)}    & 48.4 (.523) & .967 (.004) & .000 (.000) & .002 (.002) & 3.8 (1.0) & 30.7 (7.3)
 \\ 
\midrule
\textbf{MGS-edit}         & 78.2 (1.38) & .925 (.003) & .037 (.008) & .098 (.007) & 44.0 (2.2) & 21.6 (0.1) \\
\textbf{MRT-edit (+MLE 0.3)}   & 138.7 (11.1) & .929 (.011) & .227 (.094) & .472 (.066) & 178.4 (43.1) & 23.2 (1.0) \\ 
\textbf{PG-edit (+MLE 0.1)}    & 103.0 (4.05) & .904 (.001) & .051 (.016) & .246 (.027) & 68.5 (8.2) & 24.5 (0.8) \\ 
\midrule 
\textbf{Human}        &-- & -- & .000   & .011      & 107.9  & --  \\
\end{tabular}
\caption{\label{tbl:completion-main-test} Text completion results (GPT-2, Wikitext-103 test set),
reported as \texttt{mean (stdev)} using 5 random seeds.
Policy gradient (PG) and minimum risk training (MRT) are stochastically mixed with MLE and
reported as (+MLE $\alpha$), with the $\alpha$ values selected based on the task loss.
Results here are with greedy decoding; see Table~\ref{tbl:sampling} in the Appendix for ancestral sampling.
}
\end{table*}

\section{Related Work}
\paragraph{Sequence-level training for NLP.} 
Sequence-level training methods based on policy gradient have been applied to several NLP tasks \citep{liu2017improved,paulus2018deep,ziegler2019finetuning}.
Related methods use policy gradient with generative adversarial networks (GAN) \citep{yu2017seqgan,demasson2019training}. 
Policy gradient methods often face training instability and sensitivity to hyper-parameters \citep{henderson2018deep}, and GAN methods under-perform maximum likelihood \citep{caccia2020language}. 

Reward augmented maximum-likelihood (RAML) \citep{norouzi2016raml} maximizes the likelihood of sequences that are sampled proportional to their rewards, which in practice relies on a sampling method designed for a specific task loss.
Our method weights parameter, rather than sequence, samples proportional to their rewards.
Minimum risk training originated in statistical machine translation \citep{och2003minimum,smith2006minimum} and was applied to 
end-to-end neural machine translation 
\citep{shen2016mrt,edunov2018classical}.
Other approaches train a greedy decoder given a learned model \citep{gu2017trainable,chen2018stable}, which is a different setting than ours.

A separate family of methods, including globally normalized models, \citep{andor2016globally,sountsov2016length}, 
energy-based models \citep{lecun2006tutorial,wang2018learning,deng2020residual}, 
unlikelihood training \citep{welleck2020neural,li2020dont}, and beam search optimization \citep{daume2005laso,wiseman2016bso}, 
incorporate sequence-level scores without reference to an external reward function.

\paragraph{Drawbacks of MLE in NLP.}
Several studies investigate drawbacks of maximum likelihood training, including label bias \citep{lafferty2001conditional,andor2016globally}, 
exposure bias \citep{daume2009searn,ross2011reduction,bengio2015scheduled}, and loss mismatch \citep{lee2020on}.
Neural machine translation models trained with maximum likelihood have been shown to exhibit decreased performance with increased beam size \citep{koehn2017six,ott2018analyzing} and a bias towards short sequences \citep{sountsov2016length,stahlberg2019nmt}, which have been attributed to label bias due to local normalization \citep{murray2018correcting}. 

In open-ended text generation, MLE-trained models have been observed to produce non-terminating sequences \citep{welleck2020consistency}, degenerate repetition \citep{holtzman2019curious,welleck2020neural}, and a mismatched unigram distribution \citep{li2020dont}.
These motivate our investigation of an alternative training procedure.

\paragraph{Black-box optimization.}
Our approach is motivated by black-box 
optimization methods, specifically those based on random search \citep{matyas1965random,rechenberg1978,back1993overview}. 
Several methods augment random search with auxiliary information \citep{hansen2011injecting,lehman2018safe,pourchot2018cemrl}. 
Related to our method are learned manifold random search \citep{sener2020learning} which requires an inner optimization to learn parameters of a search manifold, and guided evolutionary strategies \citep{maheswaranathan2019guided} which uses surrogate directions to modify the search distribution's covariance; their method requires QR decomposition and was evaluated on synthetic and unrolled optimization tasks with smaller networks than those we consider.

\section{Experiments}
\subsection{Text Completion with GPT-2}
First, we evaluate MGS on a text completion task, which has previously been used to evaluate the effectiveness of sequence models (e.g. 
\citet{sutskever2011generating,graves2013generating,radford2018gpt2,holtzman2019curious,welleck2020neural}). 
The task consists of decoding a continuation $\hat{Y}= \mathcal{F}(\theta, X)$ given a prefix $X=(x_1,\ldots,x_k)$.

In this task, neural language models such as GPT-2 \citep{radford2018gpt2} exhibit degenerate repetition \citep{holtzman2019curious} and non-termination with greedy decoding;
\cite{welleck2020consistency} conjectured that the lack of a decoding algorithm in maximum-likelihood training is the cause of the latter.
We evaluate whether MGS, which uses a decoding algorithm during training, can alleviate these issues.

\begin{table*}[t]
\centering
\resizebox{\linewidth}{!}{
\begin{tabular}{lp{1.05\linewidth}}
\textbf{Prefix} & \textit{The British organized an expedition in early 1776 for}\\ \hdashline[0.5pt/3pt]
\textbf{MLE}    & the defense of the French colonies in the Caribbean. The expedition was led by Captain William Henry St. Clair, who had been appointed to command the expedition. The expedition was led by Captain William Henry St. Clair, who had been appointed to command the expedition in 1776. The expedition   $\rightarrow \infty$ \\\hdashline[0.5pt/3pt]
\textbf{PG-LM} &   the Continental Army. \eos\\\hdashline[0.5pt/3pt]
\textbf{PG-edit} & the defence of the French colony of New France. The expedition was to be led by Captain Samuel Frederick Gray, who was to become the first commander of the expedition. \eos  \\ \hdashline[0.5pt/3pt]
\textbf{MGS-LM} &   the defense of the Ohio River. The expedition was led by Colonel John C. St. Clair, who had been appointed to command the expedition. \eos\\\hdashline[0.5pt/3pt]
\textbf{MGS-edit} & the defense of the French colonies. The expedition was led by Lieutenant Colonel John Henry, who was promoted to lieutenant colonel in 1776. \eos  \\ 
\midrule
\textbf{Prefix} & \textit{On The Joshua Tree Tour, " Where the Streets}\\ \hdashline[0.5pt/3pt]
\textbf{MLE}    & Have No Name " was performed by the band in a medley with `` The Way You Make Me Feel ", `` The Way You Make Me Feel ", `` The Way You Make Me Feel ", and `` The Way You Make Me Feel ". The medley was included on the DVD release of the tour. \eos \\\hdashline[0.5pt/3pt]
\textbf{PG-LM} &  Have No Name " \eos \\\hdashline[0.5pt/3pt]
\textbf{PG-edit} & Have No Name " was performed by the band. The performance was recorded in the studio in Los Angeles, California, and was recorded in the same room as the band's previous two singles, `` The Way " and `` The Way ". \eos  \\ \hdashline[0.5pt/3pt]
\textbf{MGS-LM} &  Have No Name " was performed by the band. \eos \\\hdashline[0.5pt/3pt]
\textbf{MGS-edit} & Have No Name " was performed by the band in a medley with`` The Day the World Gets 'Round ". \eos  \\ 
\end{tabular}
}
\caption{Example greedy continuations (GPT-2, Wikitext-103 validation set).}
\label{tbl:continuations}
\end{table*}

\paragraph{Experimental setup.} We use the Wikitext-103 dataset \citep{merity2016pointer}, a large-scale collection of
Wikipedia articles containing over 100 million words 
that has been used for language modeling \citep{baevski2018adaptive} and text completion \citep{welleck2020neural}. 
We model individual sequences by splitting the corpus according to its newline boundaries, then splitting each sequence into a context $X$ and continuation $Y$, resulting in a dataset of $(X, Y)$ pairs.
Each continuation ends in a special $\eos$ token.
We use a context size of $k=10$ tokens, discarding sequences that are length $k$ or shorter.
The resulting dataset consists of 874,556 training, 1,896 validation, and 2,162 test pairs.

We use GPT-2 117M \citep{radford2018gpt2}, a transformer \citep{vaswani2017attention} language model with a byte-level BPE vocabulary of 50k tokens, pre-trained with maximum likelihood on WebText, a dataset of scraped web pages (see \citet{radford2018gpt2} for details).
We fine-tune the pretrained GPT-2 model using MLE and select the model state with the lowest validation perplexity.
We then continue with MGS beginning at the selected model state.
We use 4 candidates, and compute training task loss with a max decoding length of 1.3 times the ground-truth length.
Models are evaluated with a max decoding length of 500 tokens.
See Appendix~\ref{apx:ssec:expr-setup} for more details.

For the MRT and PG baselines we finetune using 4 samples. 
For policy gradient we used an exponential moving average baseline.
Each method is stochastically mixed with MLE according to a hyper-parameter $\alpha\in [0,1]$: given a training batch, we draw $z\sim \mathrm{Bernoulli}(\alpha)$ and use MLE when $z$ is zero. 
We performed a grid search using $\alpha\in \{0.1,0.3,0.5\}$, selecting $\alpha$ based on the validation task loss that the model is optimizing.
In Appendix~\ref{apx:ssec:results} we also report results for MRT and PG without stochastically mixing MLE and an ablation of the choice of MRT candidates.

Our main results are reported with greedy decoding; refer to Table~\ref{tbl:sampling} in the Appendix for results with ancestral sampling.

\paragraph{Task losses.}
We experiment with two sequence-level task losses.
We define a language modeling (\textbf{LM}) loss which scores each sequence with a fixed language model: 
\begin{align}
    c_{\text{LM}}(\hat{Y}) = -\log p_{\text{score}}(\hat{Y}).
\end{align}
Intuitively, minimizing this loss adjusts the MLE model to work well with greedy decoding.
We use the fine-tuned GPT-2 model as $p_{\text{score}}$, which is the starting point of MGS training. 
As a task loss that incorporates the ground-truth sequence, we use \textbf{edit} distance $c_{\text{edit}}(\hat{Y}, Y)$,
normalized by $|Y|$.

\paragraph{Metrics.} 
Motivated by prior work which showed that MLE-trained LMs produce repetitive, non-terminating text with greedy decoding,
we measure the portion of duplicate n-grams (we use $n=4$) \citep{welleck2020neural}
and the proportion of non-terminating continuations \citep{welleck2020consistency}:
\begin{align*}
    \textbf{repetition}(\hat{Y}) &= 1 - |\text{unique n-grams}|\big/|\text{n-grams}|,\\
    \textbf{nonterm}(\hat{Y}) &= \mathbb{I}\left[\eos\not\in \hat{Y}\right].
\end{align*}
We also report the task loss, average length of the generated continuations, and the perplexity.

\paragraph{Effect on sequence-level task loss.} Table \ref{tbl:completion-main-test} shows the task losses and metrics for the baseline fine-tuned model (MLE) and each model trained with MGS to optimize the indicated task loss (MGS-\textit{loss}).
The baseline has the highest task losses, and a high degree of non-termination (.387) and repetition (.538).
MGS-LM substantially reduces the LM task loss (59.1), along with non-termination (.012) and repetition (.035).

\begin{figure}
\begin{minipage}[t]{.49\linewidth}
\centering\includegraphics[width=1.0\columnwidth]{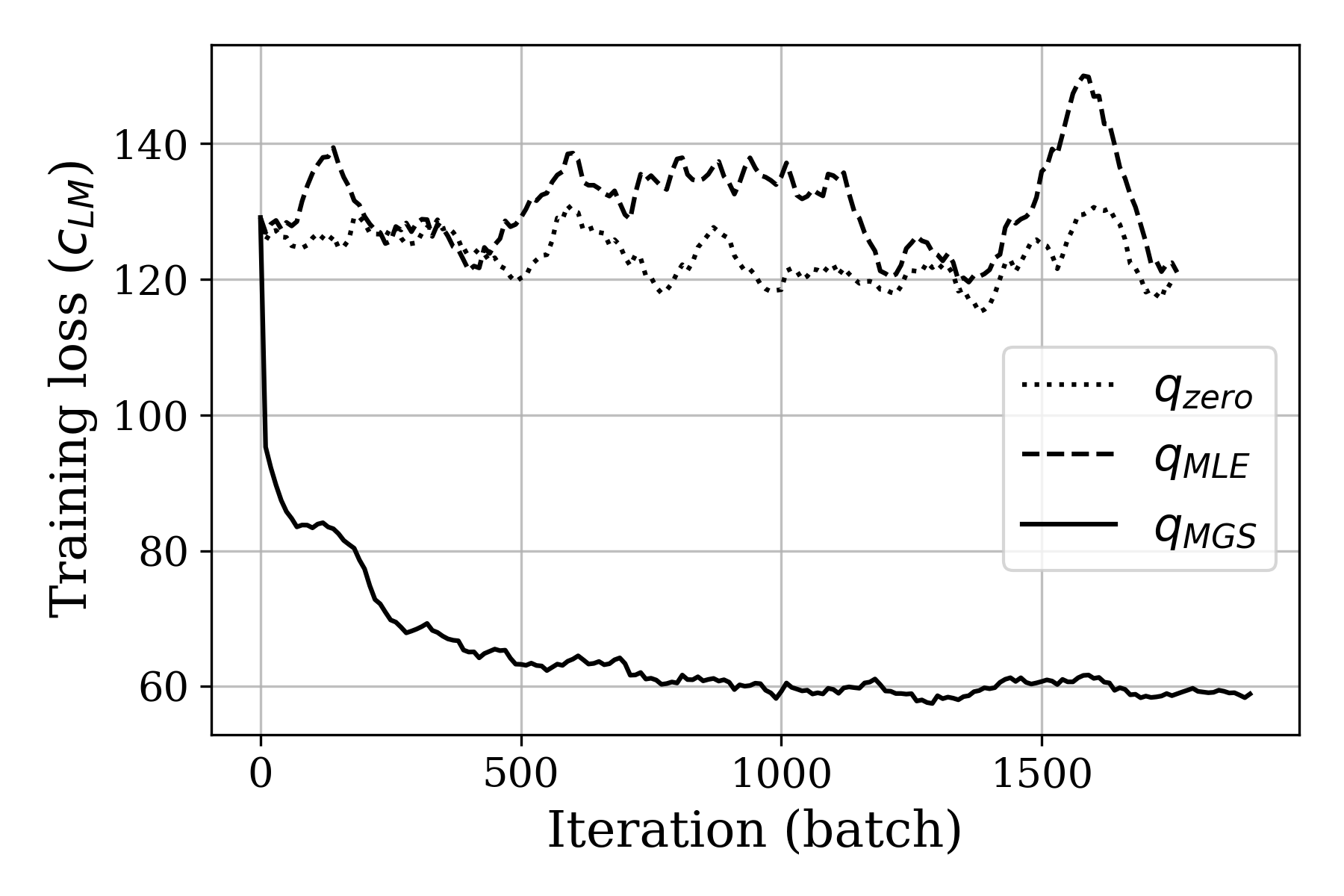}
\caption{Training task-loss ($c_{\text{LM}}$) with varied proposals.}
\label{fig:training}
\end{minipage}
\hfill
\begin{minipage}[t]{.49\linewidth}
\centering\includegraphics[width=1.0\linewidth]{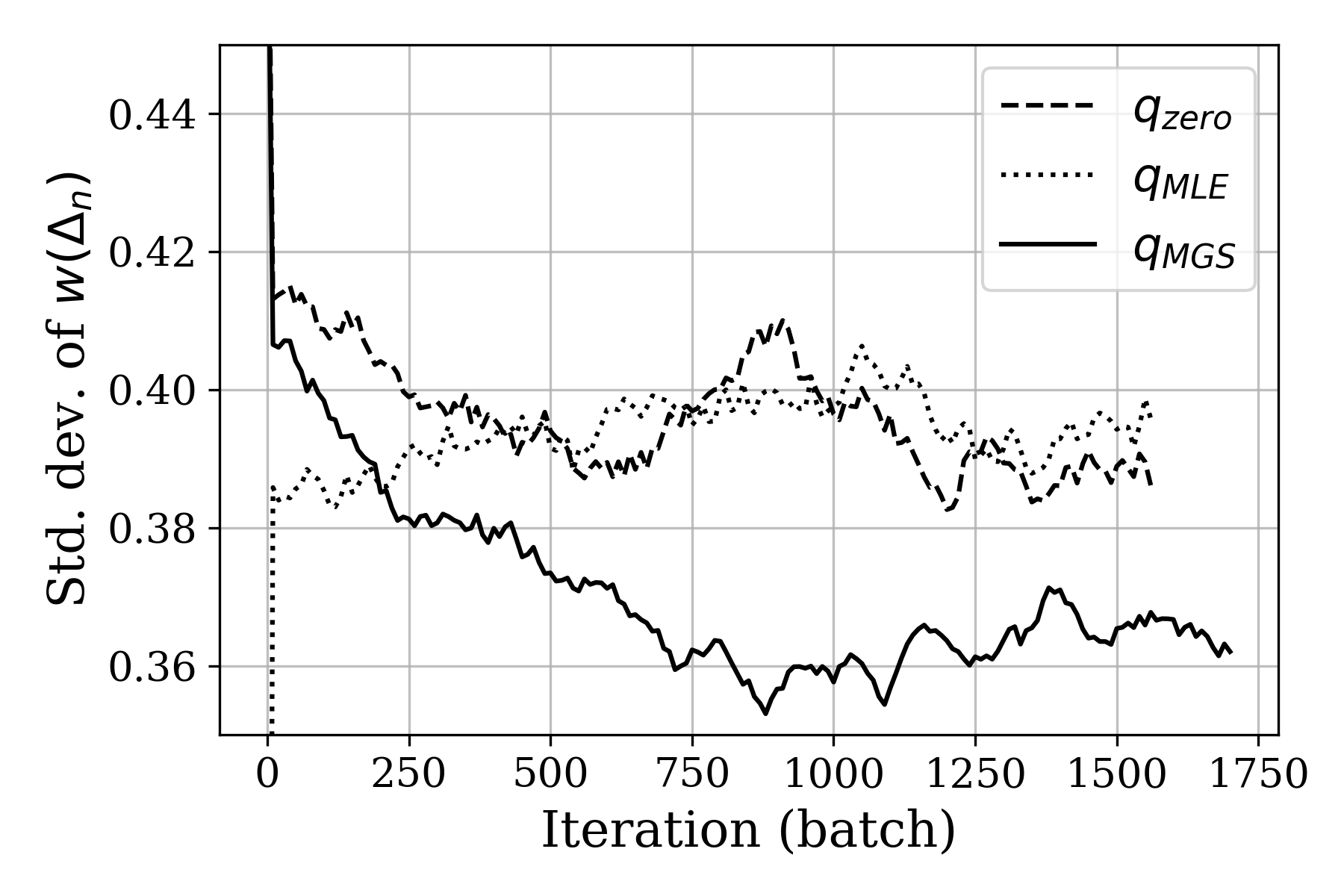}
\captionof{figure}{Standard dev. of candidate weights (MGS-LM).}
\label{fig:sd-weights}
\end{minipage}
\end{figure}

Figure \ref{fig:training} ($q_{\text{MGS}}$) illustrates how optimization progresses, with a monotonic decrease in training loss over time.
MGS-edit achieves the lowest edit distance (.928), while also substantially reducing LM task loss, non-termination, and repetition. 
Both MGS variants result in short sequences, especially MGS-LM, which is expected due to the bias towards short sequences in MLE-trained LMs \citep{stahlberg2019nmt}.

Table \ref{tbl:continuations} shows representative continuations (see the Appendix for more).
The first example shows how MGS can fix non-termination, and the second shows how MGS reduces repetition in a terminating sequence.

\begin{figure}
\includegraphics[width=\columnwidth]{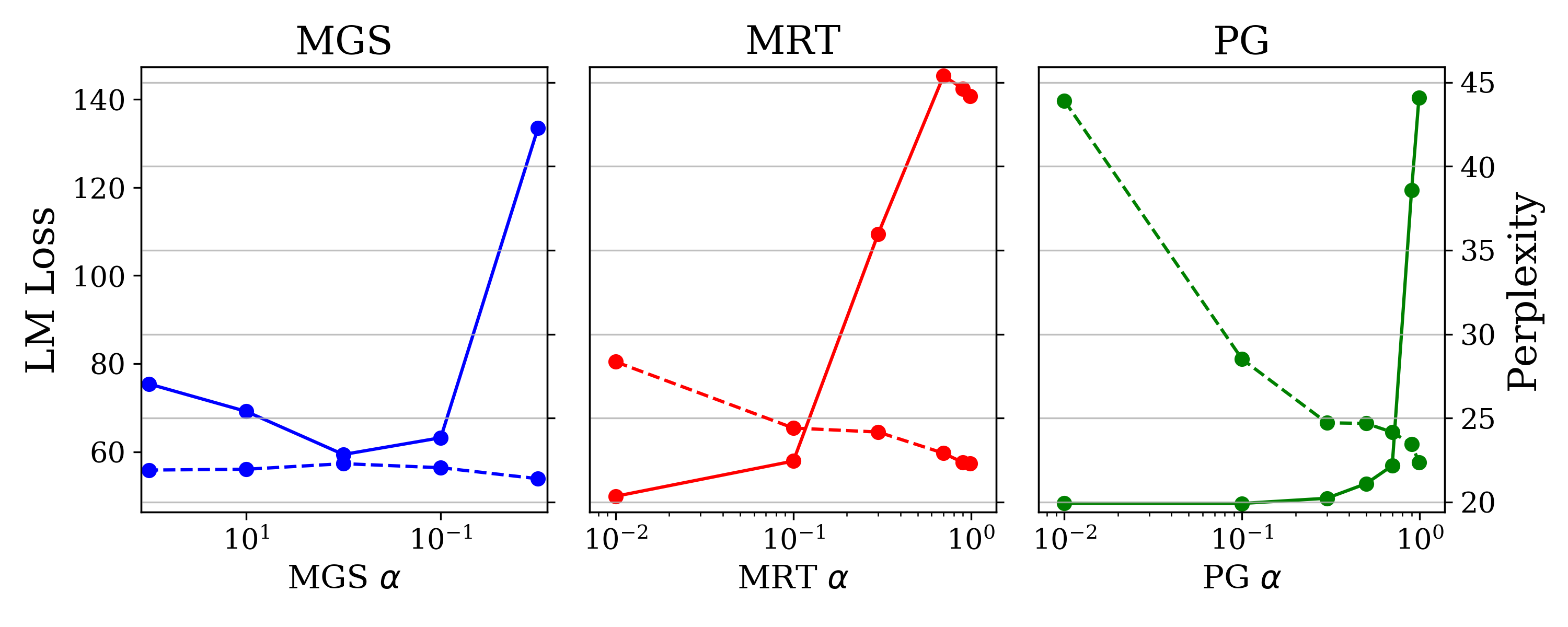}
\caption{Task loss (solid) and perplexity (dashed) as $\alpha$ varies.}
\label{fig:mrt-pg-alpha}
\end{figure}

\begin{table*}[t]
\centering
\resizebox{\linewidth}{!}{
\begin{tabular}{lp{1.05\linewidth}r}
\textbf{Prefix} & \textit{The manga was licensed for English language release by Del} & \text{Pooled Task-Loss}\\
\textbf{$\mathcal{N}_{\text{MLE}}$} & Rey in the United States, and was released in the United Kingdom in the 
  United States in the first volume of the series, and in the United States 
  in the second, and third, volumes of the series, in the United States in 
  the first and second volumes of the first and second volumes of the second  
  and third volumes ... 
  & 137.8\\
\textbf{$\mathcal{N}_{0}$} & Rey Manga in the United States. \eos & 51.2\\
\midrule
\textbf{Source} & \textit{bei den budgets der bundesstaaten geht es um sehr , sehr viel geld – ich werde ihnen die zahlen zeigen – und man kümmert sich sehr wenig um sie .} & \text{Pooled Task-Loss}\\
\textbf{$\mathcal{N}_{\text{MLE}}$} & state budgets are very , very high money -- i'll show them numbers -- and they take care of them very little . & .6767\\
\textbf{$\mathcal{N}_{0}$} & the state budgets are about a lot , a lot of money -- i'm going to show you the numbers -- and you're very little concerned about them . & .5972\\
\end{tabular}
}
\caption{Example sequences decoded from sampled candidates, showing the component that the candidate was sampled from, and the pooled cost.
Top: text completion. Bottom: machine translation.}

\label{tbl:candidates}
\end{table*}

\paragraph{PG \& MRT comparison.}
The MRT-LM and PG-LM methods result in a lower LM loss than MLE and MGS-LM.
However, the perplexity is higher than that of MGS-LM (25.8 and 30.7 vs. 22.0), with a larger standard deviation (1.7 and 7.3 vs. 0.1). 
Policy gradient finds a solution with very short sequences (average length 3.8).
For edit distance, MRT-edit underperforms MGS-edit on average (.929 vs .925), with higher nontermination, repetition, and perplexity.
PG-edit achieves the best edit distance, though with higher repetition (.246 vs. .098) and perplexity (24.5 vs. 21.6) than MGS.

We also report results with ancestral sampling in Appendix Table~\ref{tbl:sampling}.
We observe similar trends - MGS performs comparably to MRT but with better perplexity, and PG finds a degenerate short-sequence solution under the LM loss.

In summary, all three methods improve the task loss, and MGS does so while having a favorable balance across the other metrics (e.g. perplexity, repetition).
We find that $\alpha$ trades perplexity for task loss minimization in PG and MRT, while MGS finds solutions that are much more stable in terms of perplexity, as shown in Figure~\ref{fig:mrt-pg-alpha}.
Our conclusion is that MGS is an attractive alternative to mixing minimum risk training and policy gradient with maximum likelihood training for the problem of text generation.

\paragraph{MGS candidate analysis.}
First, we perform an ablation of the proposal distribution $q_{\text{MGS}}$, which is a mixture of two components.
We compare against only using the zero-mean ($q_{\text{zero}}$) or MLE-mean ($q_{\text{MLE}}$) components as proposals.
Figure \ref{fig:training} shows the training loss using each proposal, indicating that both components in the $q_{\text{MGS}}$ mixture are necessary. 
The task loss on the validation set (see Appendix Table \ref{tbl:ablation-component}) is analogous.

Next, we inspect how the pooled task loss varies between the sampled candidates.
Figure \ref{fig:sd-weights} shows the standard deviation in candidate weights $w(\Delta_k)$ during training. 
They fall within 0.35-0.45, implying that each proposal samples candidates with varied task losses. 
As a qualitative example, we sample two candidates from $q_{\text{MGS}}$ at the end of training, decode a batch of sequences with each candidate, and in Table \ref{tbl:candidates} show an example sequence and the pooled loss. 
The MLE candidate's sequence is non-terminating, while the zero candidate decodes a shorter sequence and has a lower pooled loss.

\begin{figure}
\begin{minipage}[t]{.48\linewidth}
\centering\includegraphics[width=1.0\linewidth]{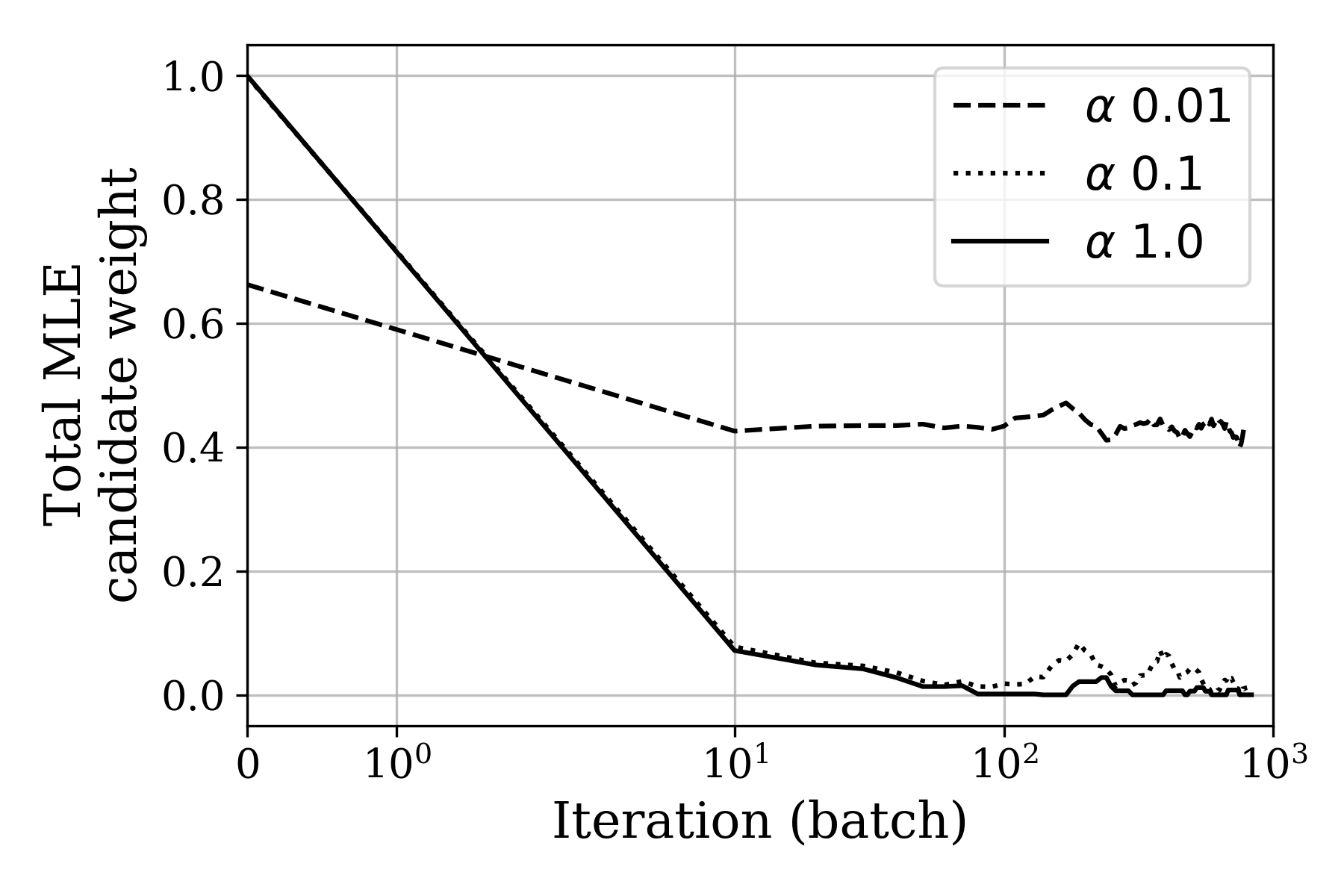}
\captionof{figure}{Total weight of candidates from the MLE component.}
\label{fig:total-weight}
\end{minipage}
\hfill
\begin{minipage}[t]{.48\linewidth}
\centering\includegraphics[width=1.0\linewidth]{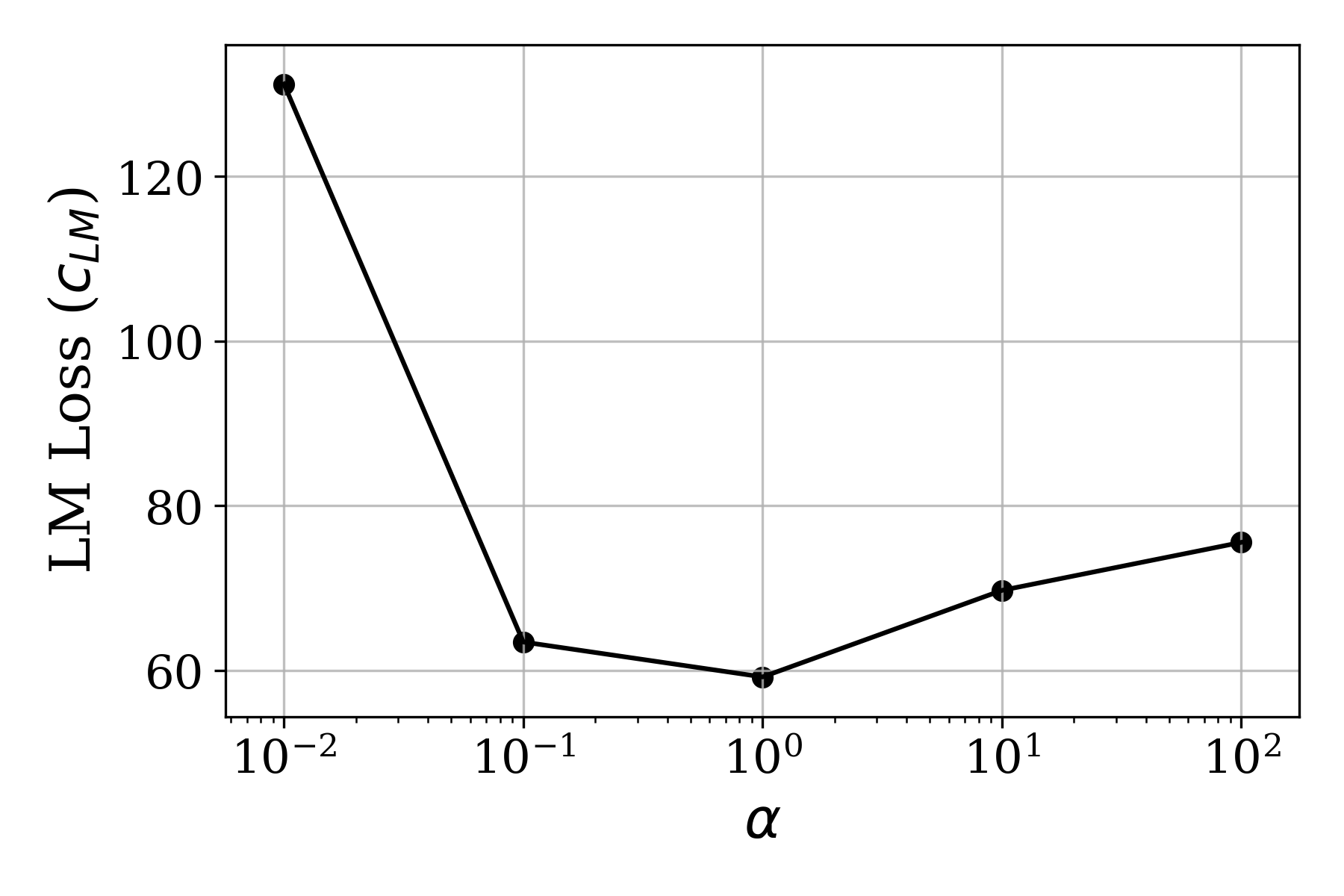}
\captionof{figure}{Validation sequence loss as $\alpha$ varies (MGS-LM).}
\label{fig:alpha}
\end{minipage}
\end{figure}

We investigate which candidates contribute to the update direction over the course of training by showing the total weight of MLE-component candidates in Figure \ref{fig:total-weight}  $(\alpha=1.0)$.
The MLE candidates are highly weighted at the beginning of training, only contributing occasionally thereafter.
Finally, we analyze the effect of the $\alpha$ hyper-parameter, which controls the entropy of the candidate weights.
As $\alpha$ decreases, the candidate weights are smoothed towards uniform, which allocates more weight to the MLE candidates, as seen in Figure \ref{fig:total-weight}.
Performance decreases when the weights are either too uniform or too peaked, as seen in Figure $\ref{fig:alpha}$.

\begin{table*}[t]
\centering
\small 
\begin{tabular}{lllllllll}
&\multicolumn{4}{c}{\textbf{Valid}} & \multicolumn{4}{c}{\textbf{Test}} \\ 
&\textsc{Bleu}$\uparrow$ & \textsc{SBleu}$\uparrow$ & \textsc{Met.}$\uparrow$ & \textsc{Edit}$\downarrow$ & \textsc{Bleu}$\uparrow$ & \textsc{SBleu}$\uparrow$ & \textsc{Met.}$\uparrow$ & \textsc{Edit}$\downarrow$  \\ 
\toprule
\textbf{MLE}                  & 36.00  & 36.22 & 63.82 & 47.88 & 34.71 & 35.67 & 62.19 & 50.74\\
\midrule
\textbf{MGS-\textsc{SBleu}}   & 36.22  & \textbf{36.58} & 64.08 & 47.25 & \textbf{35.03} & 35.89 & 62.2 & 50.23\\
\textbf{MGS-\textsc{Meteor}}  & \textbf{36.26}  & 36.51 & \textbf{64.13} & 47.35 & 34.98 & \textbf{35.97} & \textbf{62.49} & 50.29 \\
\textbf{MGS-\textsc{Edit}}    & 35.73  & 36.42 & 63.73 & \textbf{46.83} & 34.73 & 35.95 & 62.04 & \textbf{49.45}\\
\midrule
\textbf{MGS-\textsc{SBleu}} (train) & 36.19 & 36.13 & 63.65 & 48.40 & 34.80 & 35.32 & 61.95 & 51.38 \\
\end{tabular}
\caption{\label{tbl:mt-results} Machine translation results (IWSLT `14 De$\rightarrow$En). 
\textsc{Bleu} is computed with beam search (width 5).
\textsc{SBleu}, \textsc{Meteor}, and \textsc{Edit} are computed with greedy decoding to match the training conditions.
}
\end{table*}

\begin{table}
\centering
\small 
\begin{tabular}{lll}
&{\textbf{Valid}} & {\textbf{Test}} \\ 
\toprule
\textbf{\citetalias{wang2020exposure}} (MLE) & - & 34.70 \\
\textbf{\citetalias{wang2020exposure}} (MRT) & - & 35.20 \\
\midrule
\textbf{\citetalias{edunov2018classical}} (MLE) & 33.11 & 32.21 \\
\textbf{\citetalias{edunov2018classical}} (MRT) & 33.55 & 32.45 \\
\end{tabular}
\caption{\label{tbl:mt-prior} IWSLT `14 De$\rightarrow$En with minimum risk ($\textsc{Bleu}$).}
\end{table}

\subsection{Machine Translation}
\paragraph{Experimental setup.} We experiment on the IWSLT `14 German to English task \citep{cettolo2014iwslt} using a standard experimental setup from the fairseq \citep{ott2019fairseq} repository which we detail in Appendix~\ref{apx:ssec:expr-setup}. 
We train the MLE baseline and a MGS models with the same hyper-parameters. 
We use 4 candidates and a grid search over noise ($\{0.01, 0.1, 1.0\}$) and $\alpha$ ($\{1.0, 10.0, 100.0\}$).
The noise is scaled by $\frac{1}{|\theta|}\|\nabla_{\theta}\mathcal{L}_{\text{MLE}}\|_1$.

For fine-tuning, we use a batch size of 16k tokens, and accumulate gradients for 4 iterations.
We select $\alpha=100.0$ and noise $1.0$ for all MGS fine-tuning based on a grid search with MGS-\textsc{SBleu}.
For training from scratch, we select $\alpha$ 1.0 and noise 1.0.
All models are selected by validation \textsc{Bleu} using beam search with width 5.

\paragraph{Results.}
Results for the baseline, MGS fine-tuned models, and models trained from scratch with MGS are in Table \ref{tbl:mt-results}, along with prior work that fine-tuned with minimum risk training in Table~\ref{tbl:mt-prior}.

The fine-tuned MGS-S\textsc{Bleu} model improves \textsc{Bleu} over the baseline MLE model (+0.32 test) at a comparable level to the improvement from fine-tuning with MRT (+0.24 and +0.50 test),
with MGS-\textsc{Meteor} showing a similar gain.
All of the fine-tuned MGS models improve the sequence-level task losses that are computed with greedy decoding (\textsc{SBleu}, \textsc{Meteor}, \textsc{Edit}), with each model achieving the best score on its associated task loss.
MGS-\textsc{Edit} shows the largest difference, underperforming on \textsc{Bleu} yet outperforming the baseline by a full point on \textsc{Edit}.

\begin{figure}
\begin{minipage}[t]{.48\linewidth}
\centering\includegraphics[width=1.0\linewidth]{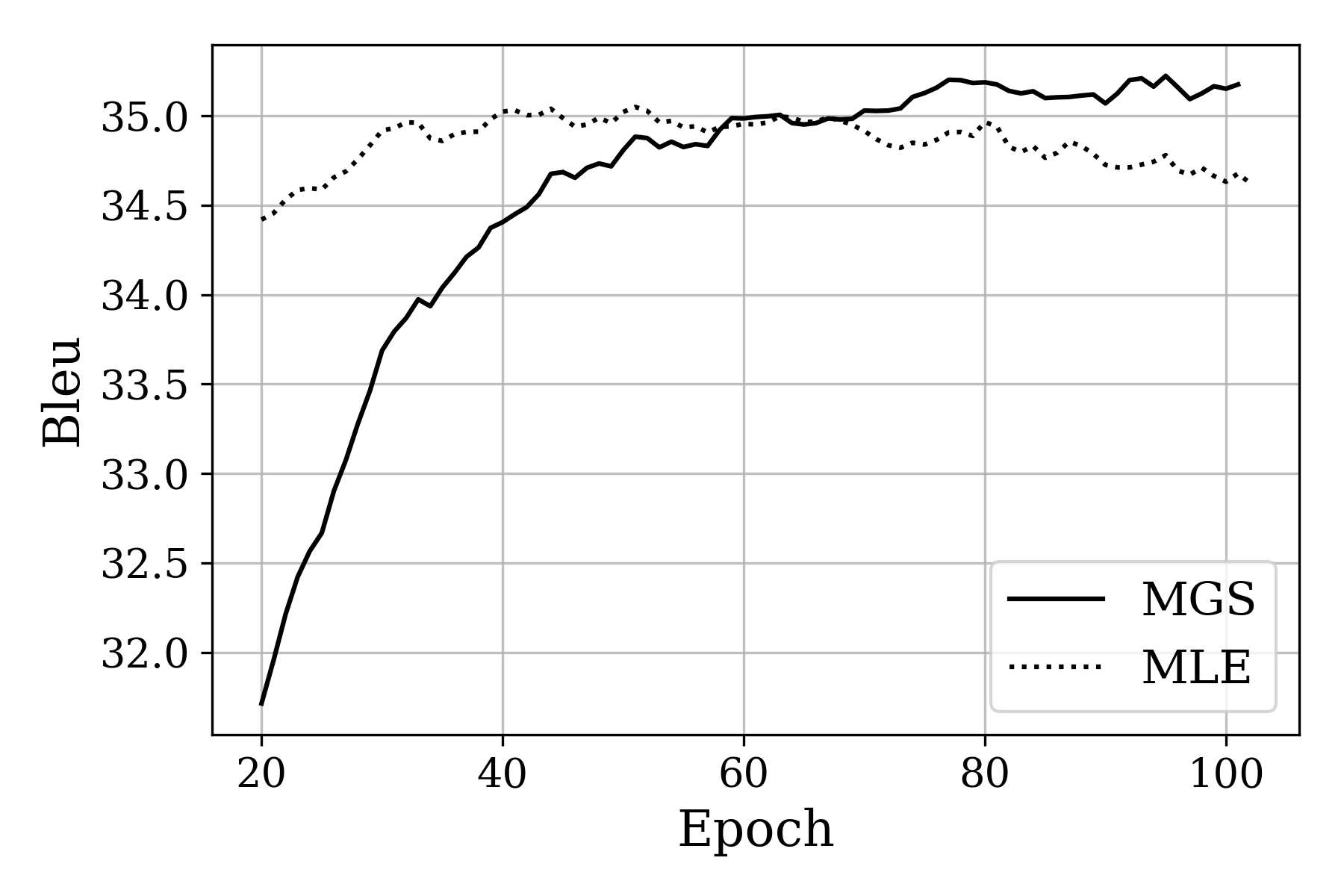}
\captionof{figure}{Validation \textsc{Bleu}.}
\label{fig:mt-bleu}
\end{minipage}
\hfill
\begin{minipage}[t]{.48\linewidth}
\centering\includegraphics[width=1.0\linewidth]{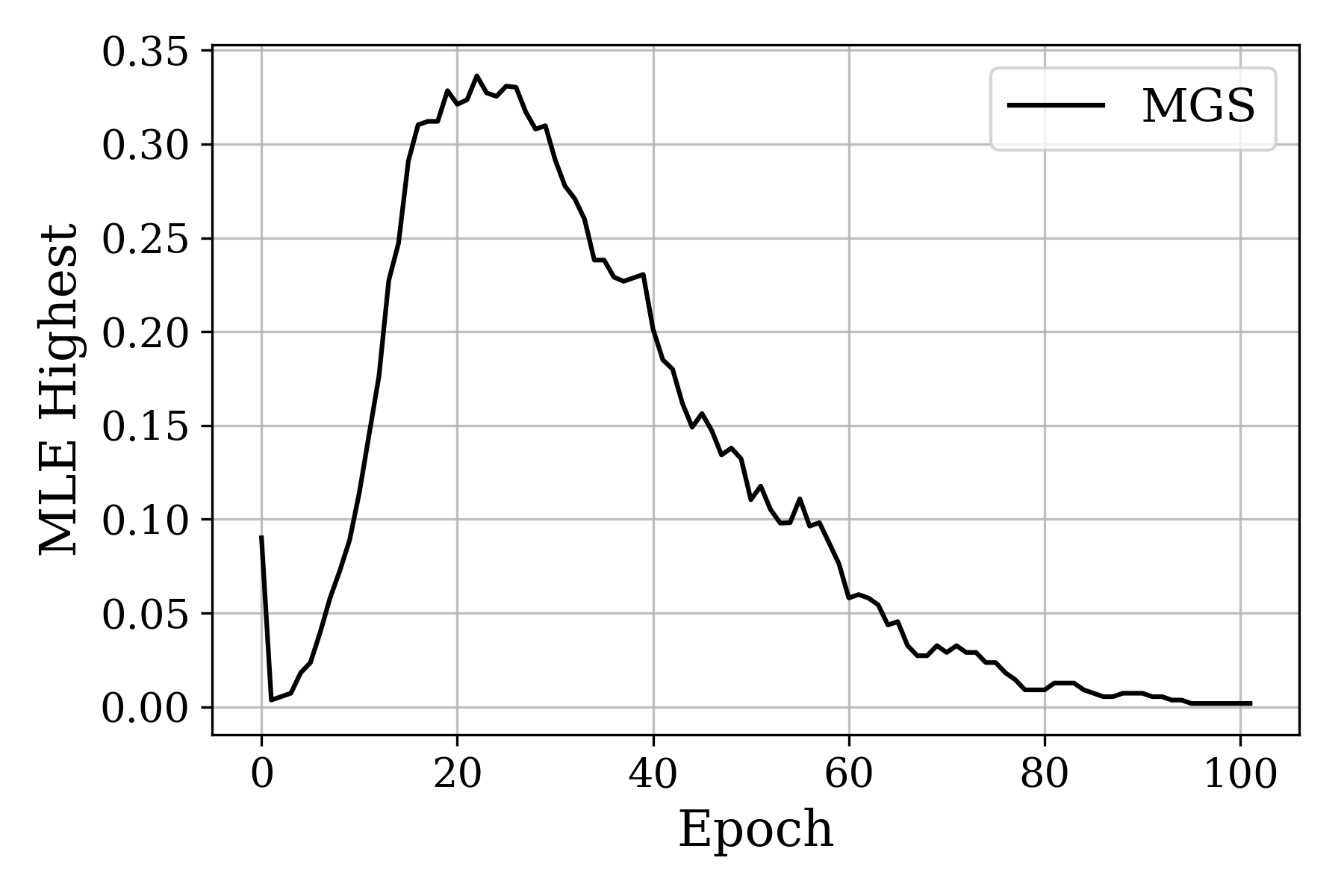}
\captionof{figure}{Proportion of highest-weight MLE candidates.}
\label{fig:mt-highest}
\end{minipage}
\end{figure}

The MGS model trained from scratch outperforms the baseline MLE model on \textsc{Bleu}, though by a smaller margin than the fine-tuned models.
Figure~\ref{fig:mt-bleu} shows the validation \textsc{Bleu} over time for MGS and the baseline, indicating that they arrive at their performance levels via different paths.
Figure~\ref{fig:mt-highest} shows the proportion of MLE candidates that had the highest weight out of the four candidates sampled from the mixture ($q_{\text{MGS}}$), and Table~\ref{tbl:candidates} shows an example sequence decoded from a candidate sampled from each component.

Candidates sampled from the zero-component tend to locally improve the task loss more than those from the MLE component.
However, we find that at the end of training, roughly 46\% of the weight comes from the MLE candidates.
We attribute this to the variations in weight between the candidates, which are smaller than those in the text completion task, with a standard deviation ranging from .005 to .025 over the course of training.

The task losses used in MT are highly concentrated on matching a reference translation and are similar to the 0-1 loss to which the log loss (MLE) is a proxy.
We suspect that it is more difficult to find candidates that improve substantially over MLE, resulting in smaller improvements than in text completion.

\section{Conclusion}
We propose maximum-likelihood guided parameter search (MGS), a training method for optimizing an arbitrary sequence-level task loss.
MGS samples update directions and weights them according to their improvement in task loss.
Key to our method is a proposal distribution which either performs random search around the current parameter or around the maximum-likelihood gradient.

MGS substantially reduced non-termination and repetition in a text completion task, and outperformed maximum likelihood on machine translation, with fine-tuning and when trained from scratch.
MGS incorporates the maximum-likelihood gradient into its objective, which led to solutions that were more stable with respect to perplexity than those found by policy and minimum risk training, which required MLE as an auxiliary loss in practice.
The results suggest that MGS is a promising alternative to minimum risk and policy gradient, and improving upon its simple, yet effective, form of exploration is a fruitful direction for future research.

\section*{Broader Impact}
Our method deals with improving neural sequence generation models for natural language processing applications, and thus inherits the potential impact and concerns that these applications bring (see \citet{brown2020gpt3} for a review). 
Generation tasks such as translation, summarization, and machine-aided writing hold the promise of improved communication, easier information access, and increased creative output, and can potentially benefit from directly optimizing task-specific objectives. 
On the other hand, generation models carry a risk of producing biased or offensive content, and can be used for nefarious applications such as fake news generation \citep{zellers2019neuralfakenews}, which could be enhanced by task loss minimization. 
Alternatively, using a task loss to specify and correct for biases in conventionally-trained models may be part of a solution that mitigates these issues, but more work is needed to determine whether this is a viable path.

\section*{Acknowledgements}
We thank Ilia Kulikov for valuable discussions.
We thank NVIDIA and eBay for their support. This work was supported by  NSF Award 1922658 NRT-HDR: FUTURE Foundations, Translation, and Responsibility for Data Science; Samsung Advanced Institute of Technology (Next Generation Deep Learning: from pattern recognition to AI); and Samsung Research (Improving Deep Learning using Latent Structure).

\bibliography{main}

\begin{thebibliography}{76}
\providecommand{\natexlab}[1]{#1}
\providecommand{\url}[1]{\texttt{#1}}
\providecommand{\urlprefix}{URL }
\expandafter\ifx\csname urlstyle\endcsname\relax
  \providecommand{\doi}[1]{doi:\discretionary{}{}{}#1}\else
  \providecommand{\doi}{doi:\discretionary{}{}{}\begingroup
  \urlstyle{rm}\Url}\fi

\bibitem[{Andor et~al.(2016)Andor, Alberti, Weiss, Severyn, Presta, Ganchev,
  Petrov, and Collins}]{andor2016globally}
Andor, D.; Alberti, C.; Weiss, D.; Severyn, A.; Presta, A.; Ganchev, K.;
  Petrov, S.; and Collins, M. 2016.
\newblock {Globally normalized transition-based neural networks}.
\newblock In \emph{54th Annual Meeting of the Association for Computational
  Linguistics, ACL 2016 - Long Papers}.
\newblock ISBN 9781510827585.
\newblock \doi{10.18653/v1/p16-1231}.

\bibitem[{B{\"{a}}ck and Schwefel(1993)}]{back1993overview}
B{\"{a}}ck, T.; and Schwefel, H.-P. 1993.
\newblock {An Overview of Evolutionary Algorithms for Parameter Optimization}.
\newblock \emph{Evolutionary Computation} ISSN 1063-6560.
\newblock \doi{10.1162/evco.1993.1.1.1}.

\bibitem[{Baevski and Auli(2019)}]{baevski2018adaptive}
Baevski, A.; and Auli, M. 2019.
\newblock Adaptive Input Representations for Neural Language Modeling.
\newblock In \emph{International Conference on Learning Representations}.
\newblock \urlprefix\url{https://openreview.net/forum?id=ByxZX20qFQ}.

\bibitem[{Bahdanau et~al.(2016)Bahdanau, Brakel, Xu, Goyal, Lowe, Pineau,
  Courville, and Bengio}]{bahdanau2016actorcritic}
Bahdanau, D.; Brakel, P.; Xu, K.; Goyal, A.; Lowe, R.; Pineau, J.; Courville,
  A.; and Bengio, Y. 2016.
\newblock An Actor-Critic Algorithm for Sequence Prediction.

\bibitem[{Bahdanau, Cho, and Bengio(2015)}]{bahdanau2014neural}
Bahdanau, D.; Cho, K.; and Bengio, Y. 2015.
\newblock Neural Machine Translation by Jointly Learning to Align and
  Translate.
\newblock In \emph{3rd International Conference on Learning Representations,
  {ICLR} 2015, San Diego, CA, USA, May 7-9, 2015, Conference Track
  Proceedings}.
\newblock \urlprefix\url{http://arxiv.org/abs/1409.0473}.

\bibitem[{Banerjee and Lavie(2005)}]{banerjee2005meteor}
Banerjee, S.; and Lavie, A. 2005.
\newblock {METEOR: An automatic metric for MT evaluation with improved
  correlation with human judgments}.
\newblock In \emph{Proceedings of the acl workshop on intrinsic and extrinsic
  evaluation measures for machine translation and/or summarization}.

\bibitem[{Bengio et~al.(2015)Bengio, Vinyals, Jaitly, and
  Shazeer}]{bengio2015scheduled}
Bengio, S.; Vinyals, O.; Jaitly, N.; and Shazeer, N. 2015.
\newblock {Scheduled sampling for sequence prediction with recurrent neural
  networks}.
\newblock In \emph{Advances in Neural Information Processing Systems}.
\newblock ISSN 10495258.

\bibitem[{Brown et~al.(2020)Brown, Mann, Ryder, Subbiah, Kaplan, Dhariwal,
  Neelakantan, Shyam, Sastry, Askell, Agarwal, Herbert-Voss, Krueger, Henighan,
  Child, Ramesh, Ziegler, Wu, Winter, Hesse, Chen, Sigler, Litwin, Gray, Chess,
  Clark, Berner, McCandlish, Radford, Sutskever, and Amodei}]{brown2020gpt3}
Brown, T.~B.; Mann, B.; Ryder, N.; Subbiah, M.; Kaplan, J.; Dhariwal, P.;
  Neelakantan, A.; Shyam, P.; Sastry, G.; Askell, A.; Agarwal, S.;
  Herbert-Voss, A.; Krueger, G.; Henighan, T.; Child, R.; Ramesh, A.; Ziegler,
  D.~M.; Wu, J.; Winter, C.; Hesse, C.; Chen, M.; Sigler, E.; Litwin, M.; Gray,
  S.; Chess, B.; Clark, J.; Berner, C.; McCandlish, S.; Radford, A.; Sutskever,
  I.; and Amodei, D. 2020.
\newblock Language Models are Few-Shot Learners .

\bibitem[{Caccia et~al.(2020)Caccia, Caccia, Fedus, {Larochelle Google Brain},
  Mila, Research, and {Canada CIFAR Chair}}]{caccia2020language}
Caccia, M.; Caccia, L.; Fedus, W.; {Larochelle Google Brain}, H.; Mila, M.;
  Research, F.~A.; and {Canada CIFAR Chair}, M.~A. 2020.
\newblock {Language GANs Falling Short}.
\newblock In \emph{International Conference on Learning Representations
  (ICLR)}.

\bibitem[{Cettolo et~al.(2014)Cettolo, Niehues, St{\"{u}}ker, Bentivogli, and
  Federico}]{cettolo2014iwslt}
Cettolo, M.; Niehues, J.; St{\"{u}}ker, S.; Bentivogli, L.; and Federico, M.
  2014.
\newblock {Report on the 11th IWSLT evaluation campaign, IWSLT 2014}.
\newblock In \emph{Proceedings of the 11th International Workshop on Spoken
  Language Translation}.

\bibitem[{Chen et~al.(2018)Chen, Li, Cho, and Bowman}]{chen2018stable}
Chen, Y.; Li, V.~O.; Cho, K.; and Bowman, S.~R. 2018.
\newblock {A stable and effective learning strategy for trainable greedy
  decoding}.
\newblock In \emph{Proceedings of the 2018 Conference on Empirical Methods in
  Natural Language Processing, EMNLP 2018}.
\newblock ISBN 9781948087841.
\newblock \doi{10.18653/v1/d18-1035}.

\bibitem[{Choshen et~al.(2020)Choshen, Fox, Aizenbud, and
  Abend}]{choshen2020weaknesses}
Choshen, L.; Fox, L.; Aizenbud, Z.; and Abend, O. 2020.
\newblock {On the Weaknesses of Reinforcement Learning for Neural Machine
  Translation}.
\newblock In \emph{International Conference on Learning Representations
  (ICLR)}.

\bibitem[{Daum{\'{e}}, Langford, and Marcu(2009)}]{daume2009searn}
Daum{\'{e}}, H.; Langford, J.; and Marcu, D. 2009.
\newblock {Search-based structured prediction}.
\newblock \emph{Machine Learning} ISSN 08856125.
\newblock \doi{10.1007/s10994-009-5106-x}.

\bibitem[{Daum\'{e} and Marcu(2005)}]{daume2005laso}
Daum\'{e}, H.; and Marcu, D. 2005.
\newblock Learning as Search Optimization: Approximate Large Margin Methods for
  Structured Prediction.
\newblock In \emph{Proceedings of the 22nd International Conference on Machine
  Learning}, ICML ’05, 169–176. New York, NY, USA: Association for
  Computing Machinery.
\newblock ISBN 1595931805.
\newblock \doi{10.1145/1102351.1102373}.
\newblock \urlprefix\url{https://doi.org/10.1145/1102351.1102373}.

\bibitem[{{De Boer} et~al.(2005){De Boer}, Kroese, Mannor, and
  Rubinstein}]{deBoer2005}
{De Boer}, P.~T.; Kroese, D.~P.; Mannor, S.; and Rubinstein, R.~Y. 2005.
\newblock {A tutorial on the cross-entropy method}.
\newblock \emph{Annals of Operations Research} ISSN 02545330.
\newblock \doi{10.1007/s10479-005-5724-z}.

\bibitem[{de~Masson~d'Autume et~al.(2019)de~Masson~d'Autume, Mohamed, Rosca,
  and Rae}]{demasson2019training}
de~Masson~d'Autume, C.; Mohamed, S.; Rosca, M.; and Rae, J. 2019.
\newblock Training Language GANs from Scratch.
\newblock In Wallach, H.; Larochelle, H.; Beygelzimer, A.; d'Alch\'{e} Buc, F.;
  Fox, E.; and Garnett, R., eds., \emph{Advances in Neural Information
  Processing Systems 32}, 4300--4311. Curran Associates, Inc.
\newblock
  \urlprefix\url{http://papers.nips.cc/paper/8682-training-language-gans-from-scratch.pdf}.

\bibitem[{Deng et~al.(2020)Deng, Bakhtin, Ott, Szlam, and
  Ranzato}]{deng2020residual}
Deng, Y.; Bakhtin, A.; Ott, M.; Szlam, A.; and Ranzato, M. 2020.
\newblock Residual Energy-Based Models for Text Generation.
\newblock In \emph{International Conference on Learning Representations}.
\newblock \urlprefix\url{https://openreview.net/forum?id=B1l4SgHKDH}.

\bibitem[{Edunov et~al.(2018)Edunov, Ott, Auli, Grangier, and
  Ranzato}]{edunov2018classical}
Edunov, S.; Ott, M.; Auli, M.; Grangier, D.; and Ranzato, M. 2018.
\newblock Classical Structured Prediction Losses for Sequence to Sequence
  Learning.
\newblock In \emph{Proceedings of the 2018 Conference of the North {A}merican
  Chapter of the Association for Computational Linguistics: Human Language
  Technologies, Volume 1 (Long Papers)}, 355--364. New Orleans, Louisiana:
  Association for Computational Linguistics.
\newblock \doi{10.18653/v1/N18-1033}.
\newblock \urlprefix\url{https://www.aclweb.org/anthology/N18-1033}.

\bibitem[{Graves(2013)}]{graves2013generating}
Graves, A. 2013.
\newblock {Generating Sequences With Recurrent Neural Networks}
  \urlprefix\url{http://arxiv.org/abs/1308.0850}.

\bibitem[{Gu, Cho, and Li(2017)}]{gu2017trainable}
Gu, J.; Cho, K.; and Li, V.~O. 2017.
\newblock {Trainable greedy decoding for neural machine translation}.
\newblock In \emph{EMNLP 2017 - Conference on Empirical Methods in Natural
  Language Processing, Proceedings}.
\newblock ISBN 9781945626838.
\newblock \doi{10.18653/v1/d17-1210}.

\bibitem[{Ha and Schmidhuber(2018)}]{ha2018world}
Ha, D.; and Schmidhuber, J. 2018.
\newblock {Recurrent world models facilitate policy evolution}.
\newblock In \emph{Advances in Neural Information Processing Systems}.
\newblock ISSN 10495258.

\bibitem[{Hansen(2011)}]{hansen2011injecting}
Hansen, N. 2011.
\newblock {Injecting External Solutions Into CMA-ES}
  \urlprefix\url{http://arxiv.org/abs/1110.4181}.

\bibitem[{Hansen and Ostermeier(2001)}]{hansen2001cmaes}
Hansen, N.; and Ostermeier, A. 2001.
\newblock {Completely derandomized self-adaptation in evolution strategies.}
\newblock \doi{10.1162/106365601750190398}.

\bibitem[{Henderson et~al.(2018)Henderson, Islam, Bachman, Pineau, Precup, and
  Meger}]{henderson2018deep}
Henderson, P.; Islam, R.; Bachman, P.; Pineau, J.; Precup, D.; and Meger, D.
  2018.
\newblock {Deep reinforcement learning that matters}.
\newblock In \emph{32nd AAAI Conference on Artificial Intelligence, AAAI 2018}.
\newblock ISBN 9781577358008.

\bibitem[{Holtzman et~al.(2019)Holtzman, Buys, Forbes, and
  Choi}]{holtzman2019curious}
Holtzman, A.; Buys, J.; Forbes, M.; and Choi, Y. 2019.
\newblock The curious case of neural text degeneration.
\newblock \emph{arXiv preprint arXiv:1904.09751} .

\bibitem[{Koehn and Knowles(2017)}]{koehn2017six}
Koehn, P.; and Knowles, R. 2017.
\newblock {Six Challenges for Neural Machine Translation}.
\newblock \doi{10.18653/v1/w17-3204}.

\bibitem[{Lafferty, McCallum, and Pereira(2001)}]{lafferty2001conditional}
Lafferty, J.; McCallum, A.; and Pereira, F. C.~N. 2001.
\newblock {Conditional random fields: Probabilistic models for segmenting and
  labeling sequence data}.
\newblock \emph{ICML '01 Proceedings of the Eighteenth International Conference
  on Machine Learning} ISSN 1750-2799.
\newblock \doi{10.1038/nprot.2006.61}.

\bibitem[{LeCun et~al.(2006)LeCun, Chopra, Hadsell, Ranzato, and
  Huang}]{lecun2006tutorial}
LeCun, Y.; Chopra, S.; Hadsell, R.; Ranzato, M.; and Huang, F. 2006.
\newblock A tutorial on energy-based learning.
\newblock \emph{Predicting structured data} 1(0).

\bibitem[{Lee et~al.(2020)Lee, Tran, Firat, and Cho}]{lee2020on}
Lee, J.; Tran, D.; Firat, O.; and Cho, K. 2020.
\newblock On the Discrepancy between Density Estimation and Sequence
  Generation.
\newblock \emph{arXiv preprint arXiv:2002.07233} .

\bibitem[{Lehman et~al.(2018)Lehman, Chen, Clune, and Stanley}]{lehman2018safe}
Lehman, J.; Chen, J.; Clune, J.; and Stanley, K.~O. 2018.
\newblock Safe Mutations for Deep and Recurrent Neural Networks through Output
  Gradients.
\newblock In \emph{Proceedings of the Genetic and Evolutionary Computation
  Conference}, GECCO ’18, 117–124. New York, NY, USA: Association for
  Computing Machinery.
\newblock ISBN 9781450356183.
\newblock \doi{10.1145/3205455.3205473}.
\newblock \urlprefix\url{https://doi.org/10.1145/3205455.3205473}.

\bibitem[{Li et~al.(2020)Li, Roller, Kulikov, Welleck, Boureau, Cho, and
  Weston}]{li2020dont}
Li, M.; Roller, S.; Kulikov, I.; Welleck, S.; Boureau, Y.-L.; Cho, K.; and
  Weston, J. 2020.
\newblock Don't Say That! Making Inconsistent Dialogue Unlikely with
  Unlikelihood Training.

\bibitem[{Liu et~al.(2017)Liu, Zhu, Ye, Guadarrama, and
  Murphy}]{liu2017improved}
Liu, S.; Zhu, Z.; Ye, N.; Guadarrama, S.; and Murphy, K. 2017.
\newblock {Improved Image Captioning via Policy Gradient optimization of
  SPIDEr}.
\newblock In \emph{Proceedings of the IEEE International Conference on Computer
  Vision}.
\newblock ISBN 9781538610329.
\newblock ISSN 15505499.
\newblock \doi{10.1109/ICCV.2017.100}.

\bibitem[{Maheswaranathan et~al.(2019)Maheswaranathan, Metz, Tucker, Choi, and
  Sohl-Dickstein}]{maheswaranathan2019guided}
Maheswaranathan, N.; Metz, L.; Tucker, G.; Choi, D.; and Sohl-Dickstein, J.
  2019.
\newblock Guided evolutionary strategies: augmenting random search with
  surrogate gradients.
\newblock In Chaudhuri, K.; and Salakhutdinov, R., eds., \emph{Proceedings of
  the 36th International Conference on Machine Learning}, volume~97 of
  \emph{Proceedings of Machine Learning Research}, 4264--4273. Long Beach,
  California, USA: PMLR.
\newblock
  \urlprefix\url{http://proceedings.mlr.press/v97/maheswaranathan19a.html}.

\bibitem[{Mania, Guy, and Recht(2018)}]{mania2018simple}
Mania, H.; Guy, A.; and Recht, B. 2018.
\newblock {Simple random search provides a competitive approach to
  reinforcement learning}.
\newblock Technical report.
\newblock \urlprefix\url{https://github.com/modestyachts/ARS.}

\bibitem[{Matyas(1965)}]{matyas1965random}
Matyas, J. 1965.
\newblock {Random Optimization}.
\newblock \emph{Automat. i Telemekh} .

\bibitem[{Merity et~al.(2016)Merity, Xiong, Bradbury, and
  Socher}]{merity2016pointer}
Merity, S.; Xiong, C.; Bradbury, J.; and Socher, R. 2016.
\newblock Pointer Sentinel Mixture Models.
\newblock \emph{ArXiv} abs/1609.07843.

\bibitem[{Murray and Chiang(2018)}]{murray2018correcting}
Murray, K.; and Chiang, D. 2018.
\newblock Correcting Length Bias in Neural Machine Translation.
\newblock In \emph{Proceedings of the Third Conference on Machine Translation:
  Research Papers}, 212--223. Brussels, Belgium: Association for Computational
  Linguistics.
\newblock \doi{10.18653/v1/W18-6322}.
\newblock \urlprefix\url{https://www.aclweb.org/anthology/W18-6322}.

\bibitem[{Norouzi et~al.(2016)Norouzi, Bengio, Chen, Jaitly, Schuster, Wu, and
  Schuurmans}]{norouzi2016raml}
Norouzi, M.; Bengio, S.; Chen, Z.; Jaitly, N.; Schuster, M.; Wu, Y.; and
  Schuurmans, D. 2016.
\newblock {Reward augmented maximum likelihood for neural structured
  prediction}.
\newblock In \emph{Advances in Neural Information Processing Systems}.
\newblock ISSN 10495258.

\bibitem[{Och(2003)}]{och2003minimum}
Och, F.~J. 2003.
\newblock {Minimum error rate training in statistical machine translation}.
\newblock \doi{10.3115/1075096.1075117}.

\bibitem[{Ott et~al.(2018)Ott, Auli, Grangier, and Ranzato}]{ott2018analyzing}
Ott, M.; Auli, M.; Grangier, D.; and Ranzato, M. 2018.
\newblock {Analyzing uncertainty in neural machine translation}.
\newblock In \emph{35th International Conference on Machine Learning, ICML
  2018}.
\newblock ISBN 9781510867963.

\bibitem[{Ott et~al.(2019{\natexlab{a}})Ott, Edunov, Baevski, Fan, Gross, Ng,
  Grangier, and Auli}]{ott2019fairseq}
Ott, M.; Edunov, S.; Baevski, A.; Fan, A.; Gross, S.; Ng, N.; Grangier, D.; and
  Auli, M. 2019{\natexlab{a}}.
\newblock fairseq: A Fast, Extensible Toolkit for Sequence Modeling.
\newblock In \emph{Proceedings of NAACL-HLT 2019: Demonstrations}.

\bibitem[{Ott et~al.(2019{\natexlab{b}})Ott, Edunov, Grangier, and
  Auli}]{ott2019scaling}
Ott, M.; Edunov, S.; Grangier, D.; and Auli, M. 2019{\natexlab{b}}.
\newblock {Scaling Neural Machine Translation}.
\newblock \doi{10.18653/v1/w18-6301}.

\bibitem[{Owen(2013)}]{owen2013monte}
Owen, A.~B. 2013.
\newblock \emph{Monte Carlo theory, methods and examples}.

\bibitem[{Papineni et~al.(2002)Papineni, Roukos, Ward, and
  Zhu}]{papineni2002bleu}
Papineni, K.; Roukos, S.; Ward, T.; and Zhu, W.-j. 2002.
\newblock {BLEU : a Method for Automatic Evaluation of Machine Translation}.
\newblock \emph{Computational Linguistics} .

\bibitem[{Paulus, Xiong, and Socher(2018)}]{paulus2018deep}
Paulus, R.; Xiong, C.; and Socher, R. 2018.
\newblock {A deep reinforced model for abstractive summarization}.
\newblock In \emph{6th International Conference on Learning Representations,
  ICLR 2018 - Conference Track Proceedings}.

\bibitem[{Plappert et~al.(2018)Plappert, Houthooft, Dhariwal, Sidor, Chen,
  Chen, Asfour, Abbeel, and Andrychowicz}]{plappert2018parameter}
Plappert, M.; Houthooft, R.; Dhariwal, P.; Sidor, S.; Chen, R.~Y.; Chen, X.;
  Asfour, T.; Abbeel, P.; and Andrychowicz, M. 2018.
\newblock {Parameter space noise for exploration}.
\newblock In \emph{6th International Conference on Learning Representations,
  ICLR 2018 - Conference Track Proceedings}.

\bibitem[{Pourchot and Sigaud(2019)}]{pourchot2018cemrl}
Pourchot; and Sigaud. 2019.
\newblock {CEM}-{RL}: Combining evolutionary and gradient-based methods for
  policy search.
\newblock In \emph{International Conference on Learning Representations}.
\newblock \urlprefix\url{https://openreview.net/forum?id=BkeU5j0ctQ}.

\bibitem[{Radford et~al.(2018)Radford, Wu, Child, Luan, Amodei, and
  Sutskever}]{radford2018gpt2}
Radford, A.; Wu, J.; Child, R.; Luan, D.; Amodei, D.; and Sutskever, I. 2018.
\newblock {Language Models are Unsupervised Multitask Learners} .

\bibitem[{Raffel et~al.(2019)Raffel, Shazeer, Roberts, Lee, Narang, Matena,
  Zhou, Li, and Liu}]{raffel2019}
Raffel, C.; Shazeer, N.; Roberts, A.; Lee, K.; Narang, S.; Matena, M.; Zhou,
  Y.; Li, W.; and Liu, P.~J. 2019.
\newblock {Exploring the Limits of Transfer Learning with a Unified
  Text-to-Text Transformer} \urlprefix\url{http://arxiv.org/abs/1910.10683}.

\bibitem[{Ranzato et~al.(2016)Ranzato, Chopra, Auli, and
  Zaremba}]{ranzato2016mixer}
Ranzato, M.; Chopra, S.; Auli, M.; and Zaremba, W. 2016.
\newblock {Sequence level training with recurrent neural networks}.
\newblock In \emph{4th International Conference on Learning Representations,
  ICLR 2016 - Conference Track Proceedings}.

\bibitem[{Rechenberg(1978)}]{rechenberg1978}
Rechenberg, I. 1978.
\newblock {Evolutionsstrategien}.
\newblock \doi{10.1007/978-3-642-81283-5_8}.

\bibitem[{Ross, Gordon, and Bagnell(2011)}]{ross2011reduction}
Ross, S.; Gordon, G.~J.; and Bagnell, J.~A. 2011.
\newblock {A reduction of imitation learning and structured prediction to
  no-regret online learning}.
\newblock In \emph{Journal of Machine Learning Research}.
\newblock ISSN 15324435.

\bibitem[{Rubinstein(1999)}]{rubinstein1999}
Rubinstein, R. 1999.
\newblock {The Cross-Entropy Method for Combinatorial and Continuous
  Optimization}.
\newblock \emph{Methodology And Computing In Applied Probability} ISSN
  1387-5841.
\newblock \doi{10.1023/A:1010091220143}.

\bibitem[{R{\"{u}}ckstie{\ss} et~al.(2010)R{\"{u}}ckstie{\ss}, Sehnke, Schaul,
  Wierstra, Sun, and Schmidhuber}]{ruckstie2010exploring}
R{\"{u}}ckstie{\ss}, T.; Sehnke, F.; Schaul, T.; Wierstra, D.; Sun, Y.; and
  Schmidhuber, J. 2010.
\newblock {Exploring Parameter Space in Reinforcement Learning}.
\newblock \emph{Paladyn, Journal of Behavioral Robotics} ISSN 2081-4836.
\newblock \doi{10.2478/s13230-010-0002-4}.

\bibitem[{Rush, Chopra, and Weston(2015)}]{rush2015}
Rush, A.~M.; Chopra, S.; and Weston, J. 2015.
\newblock {A neural attention model for sentence summarization}.
\newblock In \emph{Conference Proceedings - EMNLP 2015: Conference on Empirical
  Methods in Natural Language Processing}.
\newblock ISBN 9781941643327.

\bibitem[{Salimans et~al.(2017)Salimans, Ho, Chen, Sidor, and
  Sutskever}]{salimans2017evolution}
Salimans, T.; Ho, J.; Chen, X.; Sidor, S.; and Sutskever, I. 2017.
\newblock Evolution Strategies as a Scalable Alternative to Reinforcement
  Learning.

\bibitem[{Sener and Koltun(2020)}]{sener2020learning}
Sener, O.; and Koltun, V. 2020.
\newblock Learning to Guide Random Search.
\newblock In \emph{International Conference on Learning Representations}.
\newblock \urlprefix\url{https://openreview.net/forum?id=B1gHokBKwS}.

\bibitem[{Shen et~al.(2016)Shen, Cheng, He, He, Wu, Sun, and Liu}]{shen2016mrt}
Shen, S.; Cheng, Y.; He, Z.; He, W.; Wu, H.; Sun, M.; and Liu, Y. 2016.
\newblock {Minimum risk training for neural machine translation}.
\newblock In \emph{54th Annual Meeting of the Association for Computational
  Linguistics, ACL 2016 - Long Papers}.
\newblock ISBN 9781510827585.
\newblock \doi{10.18653/v1/p16-1159}.

\bibitem[{Smith and Eisner(2006)}]{smith2006minimum}
Smith, D.~A.; and Eisner, J. 2006.
\newblock {Minimum risk annealing for training log-linear models}.
\newblock \doi{10.3115/1273073.1273174}.

\bibitem[{Sountsov and Sarawagi(2016)}]{sountsov2016length}
Sountsov, P.; and Sarawagi, S. 2016.
\newblock {Length bias in encoder decoder models and a case for global
  conditioning}.
\newblock In \emph{EMNLP 2016 - Conference on Empirical Methods in Natural
  Language Processing, Proceedings}.
\newblock ISBN 9781945626258.
\newblock \doi{10.18653/v1/d16-1158}.

\bibitem[{Stahlberg and Byrne(2019)}]{stahlberg2019nmt}
Stahlberg, F.; and Byrne, B. 2019.
\newblock On {NMT} Search Errors and Model Errors: Cat Got Your Tongue?
\newblock In \emph{Proceedings of the 2019 Conference on Empirical Methods in
  Natural Language Processing and the 9th International Joint Conference on
  Natural Language Processing (EMNLP-IJCNLP)}, 3354--3360. Hong Kong, China:
  Association for Computational Linguistics.
\newblock \doi{10.18653/v1/D19-1331}.
\newblock \urlprefix\url{https://www.aclweb.org/anthology/D19-1331}.

\bibitem[{Sutskever, Martens, and Hinton(2011)}]{sutskever2011generating}
Sutskever, I.; Martens, J.; and Hinton, G. 2011.
\newblock {Generating text with recurrent neural networks}.
\newblock In \emph{Proceedings of the 28th International Conference on Machine
  Learning, ICML 2011}.
\newblock ISBN 9781450306195.

\bibitem[{Vaswani et~al.(2017)Vaswani, Shazeer, Parmar, Uszkoreit, Jones,
  Gomez, Kaiser, and Polosukhin}]{vaswani2017attention}
Vaswani, A.; Shazeer, N.; Parmar, N.; Uszkoreit, J.; Jones, L.; Gomez, A.~N.;
  Kaiser, {\L}.; and Polosukhin, I. 2017.
\newblock {Attention is all you need}.
\newblock In \emph{Advances in Neural Information Processing Systems}.
\newblock ISSN 10495258.

\bibitem[{Vemula, Sun, and Bagnell(2019)}]{vemula19contrasting}
Vemula, A.; Sun, W.; and Bagnell, J. 2019.
\newblock Contrasting Exploration in Parameter and Action Space: A Zeroth-Order
  Optimization Perspective.
\newblock In Chaudhuri, K.; and Sugiyama, M., eds., \emph{Proceedings of
  Machine Learning Research}, volume~89 of \emph{Proceedings of Machine
  Learning Research}, 2926--2935. PMLR.
\newblock \urlprefix\url{http://proceedings.mlr.press/v89/vemula19a.html}.

\bibitem[{Vinyals, Quoc, and Le(2015)}]{vinyals2015neural}
Vinyals, O.; Quoc, G.; and Le, V. 2015.
\newblock {A Neural Conversational Model}.
\newblock In \emph{ICML Deep Learning Workshop}.

\bibitem[{Wang and Ou(2018)}]{wang2018learning}
Wang, B.; and Ou, Z. 2018.
\newblock {Learning Neural Trans-Dimensional Random Field Language Models with
  Noise-Contrastive Estimation}.
\newblock In \emph{ICASSP, IEEE International Conference on Acoustics, Speech
  and Signal Processing - Proceedings}.
\newblock ISBN 9781538646588.
\newblock ISSN 15206149.
\newblock \doi{10.1109/ICASSP.2018.8461813}.

\bibitem[{Wang and Sennrich(2020)}]{wang2020exposure}
Wang, C.; and Sennrich, R. 2020.
\newblock {On Exposure Bias, Hallucination and Domain Shift in Neural Machine
  Translation} \urlprefix\url{http://arxiv.org/abs/2005.03642}.

\bibitem[{Welleck et~al.(2020{\natexlab{a}})Welleck, Kulikov, Kim, Pang, and
  Cho}]{welleck2020consistency}
Welleck, S.; Kulikov, I.; Kim, J.; Pang, R.~Y.; and Cho, K. 2020{\natexlab{a}}.
\newblock Consistency of a Recurrent Language Model With Respect to Incomplete
  Decoding.
\newblock \emph{arXiv preprint arXiv:2002.02492} .

\bibitem[{Welleck et~al.(2020{\natexlab{b}})Welleck, Kulikov, Roller, Dinan,
  Cho, and Weston}]{welleck2020neural}
Welleck, S.; Kulikov, I.; Roller, S.; Dinan, E.; Cho, K.; and Weston, J.
  2020{\natexlab{b}}.
\newblock Neural Text Generation With Unlikelihood Training.
\newblock In \emph{International Conference on Learning Representations}.
\newblock \urlprefix\url{https://openreview.net/forum?id=SJeYe0NtvH}.

\bibitem[{Wierstra et~al.(2014)Wierstra, Schaul, Glasmachers, Sun, Peters, and
  Schmidhuber}]{wierstra2014natural}
Wierstra, D.; Schaul, T.; Glasmachers, T.; Sun, Y.; Peters, J.; and
  Schmidhuber, J. 2014.
\newblock {Natural evolution strategies}.
\newblock \emph{Journal of Machine Learning Research} ISSN 15337928.

\bibitem[{Williams(1992)}]{williams1992}
Williams, R.~J. 1992.
\newblock {Simple statistical gradient-following algorithms for connectionist
  reinforcement learning}.
\newblock \emph{Machine Learning} ISSN 0885-6125.
\newblock \doi{10.1007/bf00992696}.

\bibitem[{Wiseman and Rush(2016)}]{wiseman2016bso}
Wiseman, S.; and Rush, A.~M. 2016.
\newblock {Sequence-to-sequence learning as beam-search optimization}.
\newblock In \emph{EMNLP 2016 - Conference on Empirical Methods in Natural
  Language Processing, Proceedings}.
\newblock ISBN 9781945626258.
\newblock \doi{10.18653/v1/d16-1137}.

\bibitem[{Wolf et~al.(2019)Wolf, Debut, Sanh, Chaumond, Delangue, Moi, Cistac,
  Rault, Louf, Funtowicz, and Brew}]{wolf2019huggingface}
Wolf, T.; Debut, L.; Sanh, V.; Chaumond, J.; Delangue, C.; Moi, A.; Cistac, P.;
  Rault, T.; Louf, R.; Funtowicz, M.; and Brew, J. 2019.
\newblock HuggingFace's Transformers: State-of-the-art Natural Language
  Processing.
\newblock \emph{ArXiv} abs/1910.03771.

\bibitem[{Yu et~al.(2017)Yu, Zhang, Wang, and Yu}]{yu2017seqgan}
Yu, L.; Zhang, W.; Wang, J.; and Yu, Y. 2017.
\newblock {SeqGAN: Sequence generative adversarial nets with policy gradient}.
\newblock In \emph{31st AAAI Conference on Artificial Intelligence, AAAI 2017}.

\bibitem[{Zellers et~al.(2019)Zellers, Holtzman, Rashkin, Bisk, Farhadi,
  Roesner, and Choi}]{zellers2019neuralfakenews}
Zellers, R.; Holtzman, A.; Rashkin, H.; Bisk, Y.; Farhadi, A.; Roesner, F.; and
  Choi, Y. 2019.
\newblock Defending Against Neural Fake News.
\newblock In Wallach, H.; Larochelle, H.; Beygelzimer, A.; d'Alch\'{e} Buc, F.;
  Fox, E.; and Garnett, R., eds., \emph{Advances in Neural Information
  Processing Systems 32}, 9054--9065. Curran Associates, Inc.
\newblock
  \urlprefix\url{http://papers.nips.cc/paper/9106-defending-against-neural-fake-news.pdf}.

\bibitem[{Ziegler et~al.(2019)Ziegler, Stiennon, Wu, Brown, Radford, Amodei,
  Christiano, and Irving}]{ziegler2019finetuning}
Ziegler, D.~M.; Stiennon, N.; Wu, J.; Brown, T.~B.; Radford, A.; Amodei, D.;
  Christiano, P.; and Irving, G. 2019.
\newblock {Fine-Tuning Language Models from Human Preferences}
  \urlprefix\url{http://arxiv.org/abs/1909.08593}.

\end{thebibliography}
\clearpage
\newpage
\appendix

\section{Appendix}

\subsection{Self-normalized Importance Sampling}
\label{apx:snis}
For completeness, we review self-normalized importance sampling (see \citep{owen2013monte} for a further review), and show the explicit derivation of the MGS update.
Importance sampling estimates the expected value of a function $f(x)$ under $p(x)$ using a proposal distribution $q(x)$.
Self-normalized importance sampling assumes $p(x)$ and $q(x)$ are only known up to multiplicative constants, $\tilde{p}(x)=ap(x)$, $\tilde{q}(x)=bq(x)$.
The expected value is estimated with weights $w(x)=\frac{\tilde{p}(x)}{\tilde{q}(x)}$,
\begin{align*}
    \tilde{\mu}_{f,p} &= \frac{\sum_{k=1}^{K}f(x_k)w(x_k)}{\sum_{k'=1}^K w(x_{k'})}\\
    &= \frac{\frac{a}{b}\sum_{k=1}^{K}f(x_k)\frac{p(x_k)}{q(x_k)}}{\frac{a}{b}\sum_{k'=1}^K \frac{p(x_k)}{q(x_k)}}\\
    &= \frac{\sum_{k=1}^{K}f(x_k)\frac{p(x_k)}{q(x_k)}}{\sum_{k'=1}^K \frac{p(x_k)}{q(x_k)}},
\end{align*}
where the last line shows that self-normalized importance sampling is equivalent to using standard importance sampling weights $\frac{p(x)}{q(x)}$ that are normalized.

In our case, $x_k$ is a direction $\Delta_k$, $f$ is the identity, and $\tilde{p}$ and $\tilde{q}$ are defined in \ref{eqn:ggs-ptilde}, \ref{eqn:proposal}.
This gives,
\begin{align*}
    w(\Delta_k) &= \frac{\exp\left(\alpha(C(\theta)-C(\theta+\Delta_n)\right)}{q_{\text{MGS}}(\Delta_n|\theta)},
\end{align*}
and 
\begin{align}
\label{apx:eqn-ggs}
    \nonumber \tilde{\Delta} &= \sum_{k=1}^K \frac{w(\Delta_k)}{\sum_{k'=1}^K w(\Delta_k)}\Delta_k \\
    &= \sum_{k=1}^K \hat{w}(\Delta_k)\exp\left(\alpha(C(\theta)-C(\theta+\Delta_k))\right)\Delta_k,
\end{align}
where
\begin{align*}
    \hat{w}(\Delta_n) &= \frac{q_{\text{MGS}}(\Delta_k|\theta)^{-1}}{\sum_{k=1}^K w(\Delta_k)}.
\end{align*}
We use the form (\ref{apx:eqn-ggs}) in Section \ref{sec:comparison}.

MGS inherits properties of self-normalized importance sampling (see \citep{owen2013monte}). 
The variance can be computed as,
\begin{align}
    \mathrm{Var}(\tilde{\Delta}_{\text{MGS}}) &= \sum_{k=1}^K \tilde{w}_k(\Delta_k - \tilde{\Delta})^2,
\end{align}
where $\tilde{w}_k = w(\Delta_k)/\sum_{k'}w(\Delta_{k'}).$ 

\subsection{Derivations}
\label{apx:mrt-derivation}
\paragraph{Minimum risk gradient.}
Consider the minimum risk training objective (\ref{eqn:mrt-cost}). 
Let $Z_{\theta}$ denote $\sum_{Y\in S}p_{\theta}(Y|X)^\alpha$, $p_{\theta}^\alpha$ denote $p_{\theta}(Y|X)^\alpha$, $q_{\theta}$ denote the distribution (\ref{eqn:mrt-cost}), and $c_{\hat{Y}}$ denote $c(\hat{Y}, Y)$.
The gradient of the objective is,
\begin{align}
\label{eqn:mrt-grad-start}
    \nabla_{\theta}\left[
        \sum_{\hat{Y}\in S}q_{\theta} c_{\hat{Y}}
    \right] &=
    \sum_{\hat{Y}\in S}q_\theta \nabla \log q_\theta c_{\hat{Y}}.
\end{align}
Now, 
\begin{align*}
    \nabla \log q_\theta &= \nabla \log p_\theta^\alpha - \nabla \log Z_\theta\\
    &= \alpha \nabla \log p_\theta - \nabla \log Z_\theta,\\
    \nabla \log Z_\theta &= \nabla\log \sum_{Y'\in S}p_{\theta}^\alpha \\
    &= \frac{\sum_{Y'\in S}\nabla p_{\theta}^\alpha}{\sum_{Y'\in S}p_{\theta}^\alpha}\\
    &= \alpha \sum_{Y'\in S}q_{\theta} \nabla\log p_{\theta}.
\end{align*}
Substituting these expressions into (\ref{eqn:mrt-grad-start}) gives $\nabla_{\theta} C_{\text{MRT}}=$
\begin{align*}
    &= 
   \sum_{\hat{Y}\in S}q_{\theta}\bigg[
     \alpha\nabla\log p_{\theta} \\ &\quad\quad\quad- \alpha \sum_{Y'\in S}q_{\theta}(Y'|X)\nabla\log p_{\theta}(Y'|X)
   \bigg]c_{\hat{Y}} \\
   \displaystyle &= \alpha \sum_{\hat{Y}\in S}q_{\theta}\bigg[
     \nabla\log p_{\theta} - \mathop{\mathbb{E}}_{Y'\sim q_{\theta}}\nabla\log p_{\theta}(Y'|X)
   \bigg]c_{\hat{Y}} \\
   \displaystyle &= \alpha\left[\mathop{\mathbb{E}}_{q_{\theta}}\left[c_{\hat{Y}}\nabla\log p_{\theta}\right]-\mathop{\mathbb{E}}_{q_{\theta}}\left[c_{\hat{Y}}\right]\mathop{\mathbb{E}}_{q_{\theta}}\left[\nabla\log p_{\theta}\right]\right],
\end{align*}
which is equation (\ref{eqn:grad-mrt}). 

When $Y^*\in S$, expanding the preceding expression for $\nabla_{\theta} C_{\text{MRT}}$ gives (hiding the conditioning terms for brevity),
\begin{align*}
    \nabla_{\theta} C_{\text{MRT}} =
    \alpha\bigg[&
        q_{\theta}(Y^*)c(Y^*)\nabla\log p_{\theta}(Y^*)+\\
        &\sum_{\hat{Y}\in S\backslash Y^*}q_{\theta}(\hat{Y})c(\hat{Y})\nabla\log p_{\theta}(\hat{Y}) -\\
        &\mathbb{E}_{q_{\theta}}\left[c_{Y^*}\right]q_{\theta}(Y^*)\nabla\log p_{\theta}(Y^*)-\\
        &\sum_{\hat{Y}\in S\backslash Y^*}\mathbb{E}_{q_{\theta}}\left[c_{\hat{Y}}\right]q_{\theta}(\hat{Y})\nabla\log p_{\theta}(\hat{Y})
    \bigg]\\
    =\alpha\bigg[& 
        w(Y^*)\nabla\log p_{\theta}(Y^*) + \\
        &\sum_{\hat{Y}\in S\backslash Y^*} w(\hat{Y})\nabla\log p_{\theta}(\hat{Y}) - \\
        &\bar{w}(Y^*)\nabla\log p_{\theta}(Y^*) -\\
        &\sum_{\hat{Y}\in S\backslash Y^*} \bar{w}(\hat{Y})\nabla\log p_{\theta}(\hat{Y})
    \bigg]\\
    =\alpha\bigg[&
        (w(Y^*)-\bar{w}(Y^*))\nabla\log p_{\theta}(Y^*)+ \\
        &\sum_{\hat{Y}\in S\backslash Y^*}(w(\hat{Y})-\bar{w}(\hat{Y}))\nabla\log p_{\theta}(\hat{Y})
        \bigg]. \qed
\end{align*}

\subsection{Limitations}
\paragraph{Computation.} Maximum-likelihood guided parameter search requires decoding $K+1$ sequences to compute the sequence costs, as well as a single forward and backward pass to compute the loss gradient $\nabla \mathcal{L}$. 
However, the candidates and their corresponding costs can be computed in parallel. 
To reduce communication cost, each parameter update can be computed by only communicating the scalar sequence costs and the random seed used to generate each perturbation, in a scheme similar to \citep{salimans2017evolution}.
In principle this would allow scaling MGS to a large number of candidate directions, which we save for future work.
We demonstrate in the experiments that MGS can also be effective with just four candidate directions computed serially.

\subsection{Experimental Setup}
\label{apx:ssec:expr-setup}
\paragraph{Text completion.}
First, we fine-tune the pretrained GPT-2 model using maximum likelihood for 400k steps, and select the model state with the lowest validation perplexity (evaluated every 5k steps).
Each training batch contains a maximum of 1024 total tokens, and we use the default hyper-parameters from the implementation in the \texttt{transformers} library \citep{wolf2019huggingface}.
We then train with MGS using the same hyper-parameters, beginning at the fine-tuned model state.
We use 4 candidates and a mixture parameter $\pi=0.5$. 
For computing each candidate's task loss during training, we use greedy decoding, with a maximum decoding length of 1.3 times the length of the longest target sequence in the batch.
The MLE gradient is clipped to have a maximum $L_2$ norm of 1.0.
The noise level $\sigma^2$ is set to $\frac{1}{|\theta|}\|\nabla_{\theta}\mathcal{L}\|_1$, which on average yields candidates with similar $L_1$ norms to the MLE gradient.
We found that scaling the noise for each weight tensor $w$ individually by $\frac{1}{|\theta_w|}\|\nabla_{\theta}\mathcal{L}\|_1$ resulted in candidates with more diverse decoded sequences, and use this method in the experiments below.
The model is evaluated on the validation set every 100 batches, and training ends when the lowest achieved validation distance does not change for 10 consecutive evaluations.

\paragraph{Machine translation.}
We experiment on the IWSLT `14 German to English task \citep{cettolo2014iwslt} using the experimental setup from the fairseq repository.\footnote{\url{https://github.com/pytorch/fairseq/tree/8e48f45aa469bbff85613520ffc161c0850e4744/examples/translation}.} 
The training data consists of 160K sentence pairs, the validation set consists of 7K sentences randomly sampled and held out from the training data, and the test data is a concatenation of \texttt{tst2010}, \texttt{tst2011}, \texttt{tst2012}, \texttt{dev2010}, and \texttt{dev2012}.\footnote{\url{https://github.com/pytorch/fairseq/blob/8e48f45aa469bbff85613520ffc161c0850e4744/examples/translation/prepare-iwslt14.sh}}
All data is lowercased and tokenized with a byte-pair encoding (BPE) of 10,000 types.
We use the \texttt{transformer\_iwslt\_de\_en} model configuration, a six-layer transformer. 
We train the MLE baseline with the default hyper-parameters, except we use gradient clipping (1.0) and disable patience (-1) which resulted in higher validation \textsc{Bleu}.
We train MGS models with the same hyper-parameters, using 4 candidates and a grid search over noise ($\{0.01, 0.1, 1.0\}$) and $\alpha$ ($\{1.0, 10.0, 100.0\}$), selecting $\alpha$ 1.0 and noise 1.0.
The noise is scaled by $\frac{1}{|\theta|}\|\nabla_{\theta}\mathcal{L}_{\text{MLE}}\|_1$.
For fine-tuning, we use a batch size of 16k tokens, and accumulate gradients for 4 iterations.
We select $\alpha=100.0$ and noise $1.0$ for all MGS fine-tuning based on a grid search with MGS-\textsc{SBleu}.
All models are selected for evaluation based on validation \textsc{Bleu} using beam search with width 5.

\subsection{Additional Results}
\label{apx:ssec:results}
\paragraph{Text completion.}
%
Table \ref{tbl:ablation-component} shows validation metrics for the proposal distribution ablation.

Table \ref{tbl:sampling} shows results using ancestral sampling.
In this case we train MGS models with ancestral sampling as the training decoder.
Evaluation is done by sampling one continuation per prefix in the validation set.

Table \ref{tbl:mrt-pg-ablation} shows results for policy gradient (PG) and minimum risk training (MRT) without mixing the MLE objective, results with mixing MLE with various mixture weights $\alpha$, and MRT with including the greedy-decoded sequence and/or the ground-truth sequence as a candidate.

Table \ref{tbl:continuations-apx} shows additional continuations.

\begin{table}
\begin{center}
\begin{tabular}{llll}
 & \textbf{Task-loss} & \textbf{Nonterm} & \textbf{PPL}\\ 
\toprule
$q_{\text{MGS}}$  & 59.2  & .013 & 22.1 \\
$q_{\text{zero}}$ & 143.0 & .348 & 20.9 \\
$q_{\text{MLE}}$  & 141.6 & .351 & 20.9 \\
\end{tabular}
\caption{Using the MGS mixture distribution ($q_{\text{MGS}}$) versus using only the zero-mean component ($q_{\text{zero}}$) or the MLE-mean component ($q_{\text{MLE}}$) as the proposal distribution in MGS ($c_{\text{LM}}$).}
\label{tbl:ablation-component}
\end{center}
\end{table}

\begin{table*}
\centering
\small 
\begin{tabular}{llllllll}
&\textbf{LM} & \textbf{Edit} & \textbf{Nonterm}  & \textbf{Repetition} & \textbf{Avg len} & \textbf{PPL}\\ 
\midrule
\textbf{MLE}  &  336.0 (27.0) & .932 (.003) & .001 (.001) & .018 (.001) & 110.5 (10.4) & 21.3 (0.2) \\
\midrule
\textbf{MGS-LM}              & 113.2 (9.1) & .934 (.003) & .000 (.000) & .006 (.001) & 28.5 (3.6) & 22.2 (0.3) \\
\textbf{MRT-LM (+MLE 0.1)}   &  117.8 (8.4) & .945 (.006) & .000 (.000) & .005 (.001) & 27.9 (3.6) & 25.8 (1.7) \\
\textbf{PG-LM (+MLE 0.1)}    &  53.2 (1.8) & .972 (.004) & .000 (.000) & .001 (.002) & 4.4 (1.0) & 30.7 (7.3) \\
\midrule
\textbf{MGS-edit} & 159.5 (8.3) & .927 (.002) & .000 (.000) & .009 (.001) & 46.0 (2.5) & 21.8 (0.2) \\
\textbf{MRT-edit (+MLE 0.3)} &  312.8 (30.8) & .930 (.004) & .002 (.004) & .021 (.004) & 105.9 (12.6) & 23.2 (1.0) \\
\textbf{PG-edit (+MLE 0.1)}  &  225.4 (16.3) & .917 (.002) & .000 (.000) & .015 (.003) & 69.2 (5.6) & 24.5 (0.8) \\
\midrule 
\textbf{Human}    &-- & -- & .000   & .009      & 107.7  & --  \\
\end{tabular}
\caption{\label{tbl:sampling} Text completion results evaluated using \textbf{ancestral sampling} 
(GPT-2, Wikitext-103 test set), 
reported as \texttt{mean (stdev)} using 5 random seeds.
The MGS models are trained with ancestral sampling as the training decoder.
}
\end{table*}

\begin{table*}
\centering
\small 
\begin{tabular}{lllllll}
& \textbf{LM} & \textbf{Edit} & \textbf{Nonterm}  & \textbf{Repetition} & \textbf{Avg. len.} & \textbf{Perplexity}\\ 
\midrule
\textbf{MLE}              & 146.4   & .938 & .379 & .545& 239.7 & 20.9 \\
\textbf{MGS-LM}           &  59.2  & .937 & .013 & .043 & 20.1 & 22.1 \\ 
\textbf{MGS-edit}         &  74.3  & .925 & .049 & .089 & 45.6 & 21.5 \\
\midrule
\textbf{MRT-LM}           & 46.8 & .968 & .000 & .000 & 3.1 & 61.9 \\ 
\textbf{MRT-LM (greedy)}  & 47.1 & .968 & .001 & .005 & 3.5 & 403.4 \\ 
\textbf{MRT-LM (gold)}    & 46.7 & .978 & .000 & .000 & 1.5 & 105719428593.4 \\ 
\textbf{MRT-LM (greedy, gold)}  & 47.3 & .980 & .000 & .000 & 1.1 & 1214304797770.7 \\ 
\textbf{PG-LM}           & 47.1 & .965 & .000 & .000 & 3.8 & 45.5  \\ 
\midrule
\textbf{MRT-LM (+MLE 0.1)}   & 55.3 & .946 & .016 & .027 & 17.9 & 24.1 \\ 
\textbf{MRT-LM (+MLE 0.3)}   & 106.6 & .911 & .065 & .218 & 78.3 & 24.0 \\ 
\textbf{MRT-LM (+MLE 0.5)}   & 135.5 & .924 & .236 & .444 & 173.9 & 22.5 \\ 
\textbf{PG-LM (+MLE 0.1)}    & 47.2 & .967 & .001 & .004 & 4.0 & 28.6 \\ 
\textbf{PG-LM (+MLE 0.3)}    & 48.2 & .960 & .004 & .009 & 7.3 & 24.5 \\ 
\textbf{PG-LM (+MLE 0.5)}    & 51.4 & .955 & .008 & .016 & 11.1 & 24.4 \\ 
\midrule
\textbf{MRT-edit}           & 159.5 & .890 & .000 & .553 & 43.6 & 34427069131.9\\ 
\textbf{MRT-edit (greedy)}  & 116.5 & .877 & .000 & .678 & 47.3 & 208.5 \\ 
\textbf{MRT-edit (gold)}    & 151.1 & .925 & .138 & .367 & 126.9 & 52.4 \\ 
\textbf{MRT-edit (greedy, gold)}  & 694.7 & .959 & .419 & .443 & 270.7 & 60.9 \\ 
\textbf{PG-edit}            & 106.5 & .895 & .016 & .196 & 57.8 & 25.6 \\ 
\midrule
\textbf{MRT-edit (+MLE 0.1)}   & 151.7 & .942 & .376 & .595 & 244.6 & 22.1 \\ 
\textbf{MRT-edit (+MLE 0.3)}   & 125.0 & .913 & .143 & .362 & 131.2 & 23.2 \\ 
\textbf{MRT-edit (+MLE 0.5)}   & 145.5 & .931 & .297 & .458 & 199.4 & 22.9 \\ 
\textbf{PG-edit (+MLE 0.1)}    & 96.4 & .901 & .016 & .183 & 47.8 & 25.4 \\ 
\textbf{PG-edit (+MLE 0.3)}    & 101.9 & .904 & .041 & .212 & 66.5 & 24.0 \\ 
\textbf{PG-edit (+MLE 0.5)}    & 109.7 & .904 & .076 & .303 & 90.9 & 24.0 \\ 
\midrule 
\textbf{Human}        &-- & -- & .000   & .009      & 107.7  & --  \\
\end{tabular}
\caption{\label{tbl:mrt-pg-ablation} Policy gradient (PG) and minimum risk training (MRT) for text completion (GPT-2, Wikitext-103 validation set, greedy decoding) in various training setups: (1)
PG and MRT without stochastic mixing of the MLE objective; (2) MRT with the greedy sequence, gold sequence, or both as candidates; (3) PG and MRT with stochastic mixing at various $\alpha$ values.
Due to the large number of combinations, we report metrics for a single random seed.
For the LM loss without MLE, MRT and PG learn to produce empty or short sequences with substantially increased perplexity.
For edit distance, MRT finds degenerate solutions in terms of either repetition, nontermination, or perplexity, even with MLE mixed in.
For MRT, no candidate strategy shows consistent improvement over the others, so for simplicity we use sampling alone in the other experiments.
Stochastically mixing in MLE with PG or MRT training controls the tradeoff between task loss and perplexity.
}
\end{table*}

\begin{table*}[t]
\centering

\resizebox{\linewidth}{!}{
\begin{tabular}{lp{1.05\linewidth}}
\textbf{Prefix} & \textit{Disappointed by her blocked entry into the operatic}\\ \hdashline[0.5pt/5pt]
\textbf{MLE}    & world, the composer turned to the theatre, where he was able to work with the theatre's theatre staff. He was able to write a number of plays, including The Mikado ( 1894 ), The Mikado ( 1894 ), The Mikado ( 1894 ), The Mikado ( 1894 ), The Mikado ( 1894 ), The Mikado ( 1894 ), The Mikado ( 1894 ), $\rightarrow\infty$ \\\hdashline[0.5pt/5pt]
\textbf{MGS-LM} & world , the composer decided to return to the theatre. \eos \\\hdashline[0.5pt/5pt]
\textbf{MGS-edit} & world , the composer was forced to return to the theatre in 1891. \eos  \\
\midrule
\textbf{Prefix} & \textit{As the 29th Brigade advanced toward the Mivo}\\ \hdashline[0.5pt/5pt]
\textbf{MLE}    & River, the 2nd Battalion, 29th Infantry Regiment, was ordered to attack the village of Mivo. The regiment was ordered to attack the village from the north, and the 2nd Battalion, 29th Infantry Regiment, was ordered to attack from the south. The regiment was ordered to attack from the north $\rightarrow\infty$ \\\hdashline[0.5pt/5pt]
\textbf{MGS-LM} & River, the brigade's commander, Brigadier General Richard H. White, ordered the brigade to move forward. The brigade's advance was halted by the arrival of the 2nd Battalion, 7th Marines. \eos \\\hdashline[0.5pt/5pt]
\textbf{MGS-edit} & River, the brigade's commander, Lieutenant Colonel Robert J. McPherson, was killed in action. \eos  \\
\midrule
\textbf{Prefix} & \textit{In April 1991, Carol Matthews and Richard Kevin Lang} \\ \hdashline[0.5pt/5pt]
\textbf{MLE}    & were hired as the new head coach of the University of Michigan. The team was ranked No. 1 in the AP Poll and No. 1 in the Coaches'Poll. The team was also ranked No. 1 in the Coaches'Poll. The team was also ranked No. 1 in the Coaches'Poll. The team was also ranked No. 1 in the Coaches'Poll. \eos\\\hdashline[0.5pt/5pt]
\textbf{MGS-LM} & were arrested for the murder of a police officer. \eos \\\hdashline[0.5pt/5pt]
\textbf{MGS-edit} & were arrested in the United States for allegedly assaulting a police officer. \eos  \\
\midrule
\textbf{Prefix} & \textit{In 1644, Hu took it upon himself to} \\ \hdashline[0.5pt/5pt]
\textbf{MLE}    & establish a new royal court in the capital, Beijing. He also established a new administrative system, which was based on the principle of the " Five Hundred Years'War ". He also established a new administrative system for the capital, which was based on the principle of the " Five Hundred Years'War ". He also established a new administrative system for the city, which was based on the principle of the " Five Hundred Years'War ". \eos\\\hdashline[0.5pt/5pt]
\textbf{MGS-LM} & establish a new capital. He built a new palace, the Palace of the Holy Apostles, and built a new church. \eos \\\hdashline[0.5pt/5pt]
\textbf{MGS-edit} & take over the administration of the province. He was assisted by the king's brother, the Duke of Wellington, who was appointed to the post of governor. \eos  \\
\end{tabular}
}
\caption{Example greedy continuations (GPT-2, Wikitext-103 validation set).
The first two show representative examples of eliminating non-termination.
Roughly 38\% of the baseline's continuations are non-terminating, with around 1\% for MGS-LM and 5\% for MGS-edit.
The next two show reduction in repetition within a terminating continuation.
}
\label{tbl:continuations-apx}
\end{table*}

\end{document}